\pdfoutput=1
\documentclass[letterpaper]{article} 
\usepackage{aaai25}
\usepackage{times}  
\usepackage{helvet}  
\usepackage{courier}  
\usepackage[hyphens]{url}  
\usepackage{graphicx} 
\usepackage{natbib}  
\usepackage{caption} 
\usepackage{algorithm}
\usepackage{algorithmic}

\usepackage{newfloat}
\usepackage{listings}
\DeclareCaptionStyle{ruled}{labelfont=normalfont,labelsep=colon,strut=off} 
\floatstyle{ruled}
\newfloat{listing}{tb}{lst}{}
\floatname{listing}{Listing}
\usepackage{tikz}
\usetikzlibrary{arrows.meta, positioning}
\usepackage{amsmath}
\usepackage{multirow}
\usepackage[table]{xcolor}
\usepackage{mdframed}
\usepackage{xcolor}
\usepackage[most]{tcolorbox}
\usepackage{etaremune}

\usepackage[utf8]{inputenc}
\usepackage{amsthm}
\usepackage{booktabs}
\usepackage{algorithm}
\usepackage{algorithmic}
\usepackage[switch]{lineno}

\usepackage{times}  
\usepackage{helvet}  
\usepackage{courier}  
\usepackage[hyphens]{url}  
\usepackage{graphicx} 
\usepackage{natbib}  
\usepackage{caption} 
\DeclareCaptionStyle{ruled}{labelfont=normalfont,labelsep=colon,strut=off} 
\usepackage{algorithm}
\usepackage{algorithmic}

\usepackage{xcolor}
\newcommand{\answerYes}[1]{\textcolor{blue}{#1}} 
 
\newcommand{\answerNA}[1]{\textcolor{gray}{#1}}

\title{Beating Hateful Narratives on Social Media with Facts: \\ Retrieval-Augmented Counter-Speech Generation}
\author{
    Greta Damo\textsuperscript{\rm 1},
    Elena Cabrio\textsuperscript{\rm 1},
    Serena Villata\textsuperscript{\rm 1}
}
\affiliations{
    \textsuperscript{\rm 1}Université Côte d'Azur, Inria, CNRS, I3S

}

\usepackage{bibentry}

\begin{document}

\maketitle

\begin{abstract}
Online hate speech threatens social cohesion by amplifying discrimination and harmful narratives against marginalized communities. While counter-speech (CS) -- fact-based, respectful responses to hate -- offers a constructive alternative to content removal, its large-scale generation remains constrained by factual reliability and the risk of generic responses. We address these challenges with a Retrieval-Augmented Generation (RAG) pipeline grounded in the first large-scale, curated knowledge base built from authoritative international and regional human rights sources, including the United Nations Digital Library, EUR-Lex, and the EU Agency for Fundamental Rights. The framework covers eight groups frequently targeted by online abuse. We systematically evaluate all pipeline components demonstrating that each design choice is empirically motivated, combining automatic metrics, LLM-based judgments, and human evaluation with comparisons against strong baselines. Results show that RAG improves factual grounding, diversity, and perceived effectiveness over No-RAG and performs favorably against existing systems. Our human evaluation further reveals differences between in-group and out-group perceptions of CS, with RAG narrowing but not eliminating this gap. These findings highlight the potential of knowledge-grounded CS for scalable social media interventions, while underscoring the need to evaluate such systems with and for the communities they are intended to serve.

\textit{{\textbf{Warning}: this paper contains explicit hateful examples some readers may find offensive.}}
\end{abstract}

%

\section{Introduction}

The rise of social media has profoundly reshaped global communication, enabling rapid information exchange and unprecedented visibility for marginalized voices \cite{lenhart2010social,ortizospina2019social,siricharoen2023social}. At the same time, these platforms have become fertile ground for the spread of \textbf{Hate Speech (HS)}. Anonymity, scale, and virality allow abusive and discriminatory narratives to proliferate with limited accountability, normalizing harmful discourse and reinforcing social divisions \cite{zimbardo1969human,mondal2017measurement,mathew2019spread}.

\begin{figure}[h]
\centering
\includegraphics[scale=0.5]{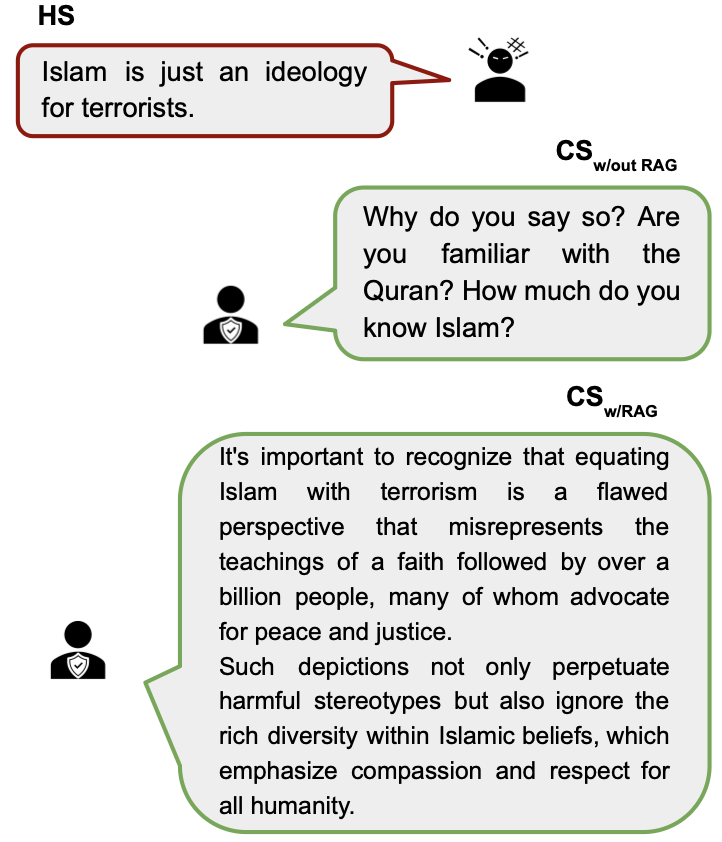}
\caption{Examples of two CS generated with GPT-4o-mini: the first without RAG, the second one with RAG.}
\label{fig:intro_ex}
\end{figure}

Platforms have responded to HS primarily through content removal, account suspension, and algorithmic downranking. These interventions are effective at reducing visibility but do little to address the underlying narratives, can be perceived as non-transparent or unevenly enforced across communities, and raise tensions with free expression \cite{coe2023}. The fragility of removal-only approaches has recently become concrete, i.e. Meta's January 2025 decision to scale back content moderation and fact-checking prompted renewed regulatory emphasis, within the EU, on platforms' obligations under the Digital Services Act and on the distinction between freedom of expression and an unqualified ``right to virality'' \cite{eu-code,meta2025}. 

\textbf{Counter-Speech (CS)} -- fact-based, respectful responses to harmful narratives -- has emerged as a constructive, rights-preserving alternative that engages with hateful content rather than erasing it \cite{benesch2014dangerous,schieb2016governing}.
Prior work shows that effective CS can reduce the persuasive impact of HS and foster more inclusive online interactions \cite{kiritchenko2021confronting}.
Yet CS's promise as a platform-level intervention is constrained by a practical challenge: manual campaigns run by NGOs and trained volunteers, while high-quality, cannot match the volume and velocity of HS circulating on social media \cite{chung2021empowering}. Systems operating at the scale of a real social media platform must therefore satisfy two conditions simultaneously: generate CS automatically and continuously, and remain trustworthy enough, i.e. factually accurate, non-generic, socially appropriate, to be shown to the targeted communities. 
Existing automatic CS systems struggle to meet both requirements. In fact, models fine-tuned on curated datasets generalize poorly to unseen HS, while unconstrained Large Language Models (LLMs) generation, despite their fluency, often lack the factual grounding needed to counter misinformation and stereotypes convincingly. Moreover, recent field evidence shows that even well-intentioned LLM-generated CS may fail to change behavior on real platforms or even backfire \cite{bar2024generative}, underscoring the need to evaluate CS for the audiences it is intended to serve, rather than through generation quality metrics alone.
Recent Retrieval-Augmented Generation (RAG) based approaches have begun to address this challenge. \citet{wilk2025fact} augment generation with web search and background knowledge, while \citet{russo2025trenteam} shows that reranker-based pipelines improve factuality and relevance. However, these systems rely on web sources that may introduce misinformation, cover a limited set of HS targets, or lack systematic analysis of retrieval and pipeline choices. No prior work has constructed a large scale, institutionally grounded knowledge base (KB) specifically for CS or systematically evaluated the contribution of its components to generation quality.
These methodological gaps point to a broader question about who such interventions actually serve. If automated CS is to operate at the scale of real platforms, its adequacy cannot be judged by text generation metrics alone, but must also be evaluated by the communities it is intended to defend, across the diversity of groups targeted by HS on online platforms.

This motivates two Research Questions (RQs): \textbf{RQ1}: \textit{can automatic CS generation be made reliably factual and non-generic enough for deployment on social media}, and \textbf{RQ2}: \textit{does the resulting CS serve the communities it targets, particularly members of those communities, who have the most at stake in how they are represented and defended}. 
We address RQ1 with a RAG pipeline for CS grounded in the first large scale, curated knowledge base built from authoritative international and regional human rights sources, i.e. the United Nations Digital Library, EUR-Lex, and the EU Agency for Fundamental Rights, eliminating the misinformation risk of web-augmented alternatives while covering eight protected groups frequently targeted by online abuse. 
We address RQ2 with a human evaluation designed not only to rate CS quality, but to compare how people who self-identify with a targeted group evaluate CS aimed at defending that group against people who do not. Figure~\ref{fig:intro_ex} shows two examples of CS, with and without RAG augmentation.

Our contributions are fourfold: \textbf{(1)} we construct the \textbf{first large-scale, authoritative KB from international and regional human rights institutions}, covering 8 protected groups across 32,792 documents, eliminating the misinformation risk of web-augmented systems 
by grounding CS exclusively in verifiable institutional sources; \textbf{(2)} we conduct a \textbf{comprehensive evaluation} of RAG-based CS, combining automatic, LLM-as-a-Judge, and human evaluations, including a comparison of in-group and out-group annotator judgments, showing that RAG narrows but does not eliminate a persistent gap in how targeted communities perceive CS; \textbf{(3)} we systematically \textbf{ablate all pipeline components} 
demonstrating that each design choice is empirically motivated for producing concise CS, deployable in real social media contexts; and \textbf{(4)} we will \textbf{release generated CS instances with their retrieved evidence and the collected KB}, supporting reproducibility and future research\footnote{ \url{https://anonymous.4open.science/r/ragcs-D789} has a data sample in CSV format, following FAIR principles (FORCE11 2020). Full materials will be released upon paper acceptance.}. 

\section{Related Work}

\paragraph{CS in Online Social Media.}
HS is a persistent social media harm that affects targeted communities and can shape the tone and dynamics of online discussions. Beyond content removal and other platform-level moderation interventions, CS has been proposed as a complementary approach that directly responds to hateful narratives by providing alternative perspectives, factual information, or support to targeted groups \cite{benesch2014dangerous,schieb2016governing}. Because manually written CS cannot scale to the volume and speed of harmful content on major platforms \cite{chung-etal-2021-towards}, research has increasingly explored automatic CS generation. Importantly, recent work shows that effectiveness in online environments cannot be inferred from linguistic quality alone. \citet{hong-etal-2024-outcome} steer generation toward conversational outcomes such as reduced incivility and non-hateful re-engagement, while \citet{bar2024generative} show in a large scale Twitter/X field experiment that LLM-generated contextualized CS can be ineffective or even backfire, in contrast to simpler interventions. Similarly, \citet{cima2025contextualized} find that adapting CS to the moderation context and individual users can improve its effectiveness. These findings highlight that CS generation is not only a language generation problem, but also an intervention problem shaped by platform context, audience, and user behavior.

\paragraph{CS Resources and Generation.}
Early CS datasets were collected from social media platforms such as YouTube, Reddit, and Gab \cite{mathew2019thou,qian2019benchmark}, with subsequent resources incorporating conversational context \cite{yu2022hate}, multiple target groups in MT-CONAN \cite{fanton2021human}, and dialogue in DIALOCONAN \cite{bonaldi2022human}. Knowledge-grounded resources and generation methods subsequently tried to make CS more specific and factually supported \cite{chung-etal-2021-towards,bonaldi2025first}. Generation approaches have evolved from sequence-to-sequence models \cite{qian2019benchmark} to retrieval-based systems \cite{zhu-bhat-2021-generate,chung-etal-2021-towards}, controllable generation \cite{bonaldi-etal-2023-weigh,saha2022countergedi,gupta2023counterspeeches}, and more recently zero- and few-shot LLM prompting \cite{ashida2022towards,tekiroglu2022using}. However, recent evaluations reveal limitations that are particularly relevant for social media deployment. \citet{ngueajio-etal-2025-think} find that LLM-generated CS can be overly verbose and inaccessible, while \citet{martone2026assisted} show that even knowledge-grounded LLM outputs frequently require expert post editing to meet quality and guideline requirements. This literature suggests that scalable CS must balance factuality and generation quality with the constraints and expectations of online audiences.


\paragraph{CS Evaluation and Knowledge Grounding.}
CS is commonly evaluated using automatic 
metrics, as well as human or LLM-based judgments. Because reference-based metrics correlate imperfectly with human assessments, recent work has introduced multi-dimensional and reference-free evaluation frameworks \cite{zhu2023judgelm,zhong-etal-2022-towards,jones-etal-2024-multi,damo2025effectiveness,hengle2025cseval}. \citet{song-etal-2025-assessing} further show that AI-generated and human-written CS exhibit systematic linguistic differences that can be detected by both classifiers and humans, raising questions about how automatic responses may be perceived by online audiences. Retrieval-augmented generation has consequently emerged as a way to ground CS in external evidence. 
\citet{chung-etal-2021-towards} retrieve relevant material from Newsroom and WikiText-103, while \citet{jiang2025rezg} retrieve counter-arguments from the ChangeMyView subreddit, thus grounding generation in arguments produced by users in an online debate community. More recent approaches use news and web-based evidence. \citet{russo2025trenteam} build on the ML-MTCONAN-KN dataset \cite{bonaldi2025first} and use the retrieval and selection procedure of \citet{chung-etal-2021-towards}, retrieving evidence from the Newsroom news corpus and WikiText-103. \citet{wilk2025fact} use Newsroom as their primary knowledge source and complement it with the Tavily web-search API to retrieve more recent documents. These approaches differ in the nature of their evidence. Community-generated arguments may better reflect the language and rhetorical conventions of social media, while news and web sources provide broader and more current factual information. Institutional sources, in contrast, offer greater authority and traceability but may be less directly aligned with how users communicate and engage online. This trade-off is particularly important for CS, where factual reliability alone does not guarantee that a response will be persuasive or acceptable to its intended audience.

Our work addresses this gap by grounding CS in a large scale KB constructed from authoritative international and regional human rights sources, while evaluating the resulting responses with human participants, including members of targeted communities. Rather than treating retrieval solely as a means to improve generation metrics, we investigate whether institutional evidence can produce concise, factually grounded, and socially appropriate CS for online interactions. This connects knowledge grounded generation with the broader question of how automated interventions are perceived by the communities they are intended to support.

\section{Knowledge Base Construction}
\label{sec:kb}

As a first step, we build a comprehensive KB designed to ensure maximal coverage of documents addressing groups commonly targeted by HS. The goal is to gather all relevant materials, such as reports, resolutions, and legal texts, that provide evidence or context regarding these targets.
Specifically, we focus on 8 target groups: Women, People of Color, Persons with Disabilities, Migrants, Muslims, Jews, LGBT+ persons, and Others. These categories align with the classic targets in the literature, including the ones of our baseline MultiTarget-CONAN \cite{fanton2021human}.
We used GPT-based prompting to generate synonyms and semantically related keywords, ensuring that queries captured diverse terminology across cultural and policy contexts (keywords and prompts are in Appendix \ref{app:keywords} and \ref{app:prompt}). To guarantee relevance, we perform a keyword-based search so that retrieved knowledge is directly aligned with the HS targeting these groups. To ensure reliability in the generated CS, we rely exclusively on institutional publicly available sources, i.e. the United Nations Digital Library, EUR-Lex, and the European Union Agency for Fundamental Rights (FRA). To construct our KB, we follow three main steps: document retrieval, PDF-to-text conversion, and knowledge base integration.



\textbf{Document Retrieval.} The \textbf{United Nations Digital Library} serves as the central repository for official UN documents on human rights, equality, and anti-discrimination. We develop a custom crawler using \texttt{requests} and \texttt{BeautifulSoup} to systematically combine three query dimensions: target keywords, document types (e.g. resolutions, treaties, NGO statements), and years (2000-2025). The crawler paginates through results, extracts and normalizes documents in English, and organizes downloads by target group. A metadata file records \texttt{id, fname, target, type, year, url}. Error handling covers duplicate checks, retries, and skips completed downloads. 
At the European level, \textbf{EUR-Lex} provides access to EU law, treaties, and legal acts, while the \textbf{EU Agency for Fundamental Rights (FRA)} publishes reports on human rights within the EU. Using the same target keywords, we retrieve documents in English (2000–2025) from both sources, following the same metadata structure as the UN corpus. These sources complement the UN materials, forming a multi-level knowledge base, global and European, that ensures comprehensive and reliable grounding for CS generation.


\textbf{PDF Text Extraction.} UN and EU documents are converted into plain text using PyMuPDF to extract machine-readable text, while Tesseract OCR handles scanned PDFs. The process is parallelized for efficiency and designed to be fault-tolerant, with outputs stored incrementally in JSON batches linked to document IDs.


\textbf{Knowledge Base Integration.} All materials from UN, EUR-Lex, and FRA are standardized into JSON with metadata (target group, document type, year, URL). The KB spans the years 2000–2025 and combines factual and policy-oriented resources by target group and document type. To the best of our knowledge, this is the first large-scale, authoritative KB specifically constructed for CS generation. 


\begin{table}[h!]
\centering
\small
\begin{tabular}{lrrrr}
\toprule
\textbf{Keyword} & \textbf{EU \#} & \textbf{EU wds} & \textbf{UN \#} & \textbf{UN wds} \\
\midrule
Disabled     & 20   & 17{,}980 & 1{,}718  & 9{,}705 \\
Human rights & 72   & 35{,}070 & 13{,}396 & 10{,}048 \\
Jews         & 29   & 25{,}111 & 178      & 5{,}707 \\
LGBT+         & 22   & 26{,}953 & 164      & 5{,}998 \\
Migrants     & 61   & 25{,}948 & 3{,}784  & 8{,}748 \\
Muslim       & 11   & 21{,}340 & 406      & 3{,}272 \\
POC          & 38   & 27{,}897 & 4{,}840  & 8{,}365 \\
Women        & 13   & 23{,}122 & 8{,}040  & 8{,}006 \\
\midrule
\textbf{Total / Avg.} & \textbf{266} & \textbf{25{,}928} & \textbf{32{,}526} & \textbf{7{,}669} \\
\bottomrule
\end{tabular}
\caption{Comparison of EU (FRA and Eur-Lex) and UN documents by keyword group.} 
\label{tab:fra-un-compact}
\end{table}

Table~\ref{tab:fra-un-compact} reports descriptive statistics of the UN and EU (EUR-Lex + FRA) corpora by target group keyword. While the UN collection is considerably larger in terms of the number of documents (over 32k), EU reports are more detailed, with substantially higher average word counts per document ($\approx 26k$ vs. $\approx 7.7k$). Coverage also varies by group: for example, the UN corpus contains a large volume of material on human rights and Women, while EU provides more in-depth analyses on Migrants, LGBT+ individuals, and People with Disabilities. Together, these complementary sources balance breadth (UN) with depth (EU), 
ensuring that our KB is both comprehensive and adaptable to different CS scenarios.

\section{Counter-Speech Generation Pipeline}

\begin{figure}[ht!] 
    \centering
    \includegraphics[width=0.48\textwidth]{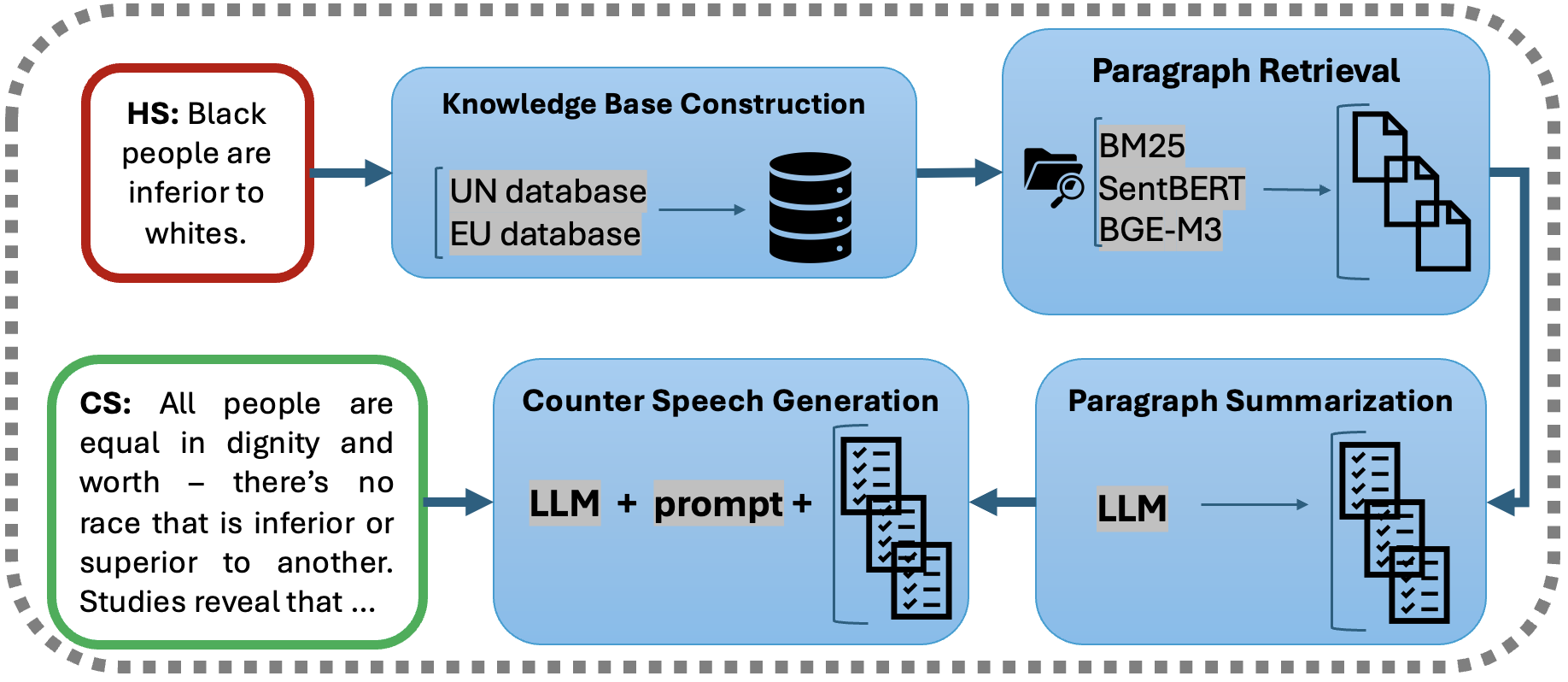} 
    \caption{ Overview of our RAG-based CS generation pipeline. Step 1: paragraph retrieval from a domain-specific KB; Step 2: LLM paragraph summarization; Step 3: CS generation conditioned on the summarized knowledge.}
    \label{fig:pipeline}
\end{figure}

Our RAG-based framework integrates three key components: (i) paragraph retrieval, (ii) paragraph summarization, and (iii) CS generation. Figure \ref{fig:pipeline} provides an overview. 

\subsection{Paragraph Retrieval}
Our domain-specific KB is segmented into paragraphs and tokenized for fine-grained access (with \texttt{gpt-2}). Each paragraph is embedded and stored using FAISS \cite{douze2024faiss} to allow efficient similarity search, obtaining a total of 3,173,630 tokenized paragraphs (details in Appendix \ref{app:prompt}) . 

Given a HS message $h$, we aim to retrieve a small set of relevant paragraphs from the KB (i.e., the most similar to the target HS).
For this, we employ three complementary retrieval models: BM25, Sentence-BERT (SBERT), and BGE-M3.  
Let the KB consist of paragraphs $\mathcal{P} = \{p_1, p_2, \ldots, p_N\}$. For each retriever 
$r \in \{\text{BM25}, \text{SBERT}, \text{BGE-M3}\} $%
, we compute a similarity score:
$s_r(h, p_i) = \text{sim}_r(\phi(h), \phi(p_i))$, 
where $\phi(\cdot)$ is the embedding (or term-weight) function induced by the retriever. For BM25, the similarity score is based on TF-IDF, and document length normalization. For SBERT and BGE-M3, similarity is measured as the cosine similarity between dense vector embeddings of the HS and candidate paragraph, capturing semantic relatedness even in the absence of exact lexical overlap.
For each retriever, we obtain the top-$k$ ranked paragraphs:
$R_r(h) = \{p_{r,1}, p_{r,2}, \ldots, p_{r,k}\}, \quad k=3.$

\noindent The choice of $k=3$ ensures high precision while providing enough external knowledge to support short and specific CS responses suited for social media. Each retrieved paragraph is stored with its source document identifier for traceability.  

\subsection{Paragraph Summarization}
Although retrieval provides relevant context, LLMs have limited context windows, making direct use of full paragraphs costly. To address this, we summarize retrieved paragraphs before passing them to the generation stage. For each retriever $r$ and each paragraph $p_{r,j} \in R_r(h)$, we obtain a summary using one out of four LLMs:
$m \in \{\text{GPT-4o-mini}, \text{Llama}, \text{Command-R}, \text{Mistral}\}$. Formally, the summarization function is:
\[
\sigma_m(p_{r,j}), 
\]
which produces a condensed version of paragraph $p_{r,j}$ w.r.t. the generation task (with \texttt{max\_new\_tokens=150} and \texttt{temperature=0.5}).  
This yields a set of summarized paragraphs:
\[
S_{r,m}(h) = \{\sigma_m(p_{r,1}), \sigma_m(p_{r,2}), \sigma_m(p_{r,3})\}.
\]

\noindent Summaries are stored in CSV format with their source IDs, ensuring alignment between retrieval and generation stages (prompts are in Appendix \ref{app:prompt}).  

\subsection{Counter-Speech Generation}
In the final stage, the summaries are used as external knowledge to generate CS. For each HS message $h$, retriever $r$, and model $m$, the generation function is defined as:
\[
c_{r,m}(h) = \text{LLM}_m(h, S_{r,m}(h)),
\]
where $c_{r,m}(h)$ denotes the CS generated by model $m$ conditioned on the HS instance $h$ and the summarized knowledge $S_{r,m}(h)$. Since we use three retrievers and four LLMs, the system produces: $|C(h)| = | \{c_{r,m}(h)\} | = 3 \times 4 = 12$ CS outputs for each HS instance, enabling systematic comparison across retrieval methods and models. 
To ensure that generated CS is deployable in (online) real-world contexts, we restrict outputs to a maximum of two sentences (\texttt{max\_new\_tokens=150} and \texttt{temperature=0.5}). This reflects social media's concise communication norms, where posts are typically 1-2 sentences long \cite{csahinucc2021tweet}, and prior work shows that concise, relatable CS is more engaging and effective against harmful narratives \cite{russo-etal-2023-countering,bonaldi2024safer,benesch2016counterspeech}.

\subsection{Models and Retrievers}
We evaluate our pipeline with four LLMs of comparable parameter scale (7–8B), and three retrievers representing both sparse and dense approaches.

\textbf{LLMs.} We use Meta-Llama-3.1-8B-Instruct\footnote{\url{https://huggingface.co/meta-llama/Llama-3.1-8B-Instruct}}, Cohere's Command-R-7B 
\cite{cohere2025commandaenterprisereadylarge}, Mistral-7B-Instruct-v0.3\footnote{\url{https://huggingface.co/mistralai/Mistral-7B-Instruct-v0.3}}, and gpt-4o-mini-2024-07-18. They are selected as LLMs of similar parameter size, balancing efficiency and accuracy, allowing comparison across open-weight and proprietary settings, and are used in related works we use for comparison.

\textbf{Retrievers.} \textbf{BM25} \cite{robertson2009probabilistic} is a sparse lexical retriever based on TF-IDF with document length normalization. It is an effective baseline for keyword-sensitive domains where exact term overlap is important. \textbf{Sentence-BERT (SBERT)} \cite{reimers2019sentence} encodes queries and passages into dense vector embeddings optimized for semantic similarity, then ranked via cosine similarity.
\textbf{BAAI General Embeddings (BGE-M3)} \cite{chen2024bge} is a recent dense embedding model trained for multilingual and multi-task semantic retrieval. 
These retrievers capture complementary dimensions of relevance, from exact term matching (BM25) to semantic similarity (SBERT, BGE-M3), ensuring robust retrieval across different types of hateful content and knowledge sources.

\section{Experimental Setup}

We experiment on the Multi-Target CONAN (MT-Co) dataset \cite{fanton2021human}, which contains 5,003 HS/CS pairs in English covering multiple target groups, including People with Disabilities, Jews, LGBT+ individuals, Muslims, Migrants, People of Color (POC), and Women. Collected through a human-in-the-loop process, MT-Co provides high-quality, contextually relevant CS, and serves as a first baseline for our experiments. Additionally, we compare our pipeline with the four instruction-tuned LLMs of comparable size without RAG, and we compare our results with competitive approaches~\cite{russo2025trenteam,wilk2025fact,bonaldi2025first}\footnote{We cannot compare to \cite{jiang2025rezg} due to lack of publicly available code and data, even upon request.}.


\section{Metrics} \label{sec:metrics}


\paragraph{Automatic Metrics.}
We evaluate the quality of generated CS using a combination of reference-based, reference-less, and LLM-based methods. 
\textbf{Reference-based Metrics.} To evaluate alignment with human-written CS in MT-Co, we report: \textbf{BLEU-4}~\cite{papineni2002bleu}, \textbf{ROUGE-L}~\cite{lin-2004-rouge}, \textbf{METEOR}~\cite{banerjee2005meteor}, and \textbf{BERTScore}~\cite{zhang2019bertscore}.
\textbf{Reference-less Metrics.} To complement reference-based evaluation, we assess intrinsic qualities of generated CS. \textbf{Distinct-1/2} \cite{li2015diversity} measures the proportion of unique unigrams and bigrams, reflecting lexical diversity. \textbf{Repetition Rate (RR)} \cite{cettolo2014repetition} measures the fraction of repeated n-grams within a generation, capturing redundancy. They are complementary as RR targets repeated n-grams specifically rather than overall lexical variety. \textbf{Safety}: OpenAI’s content moderation API scores each output across categories of potential harm (e.g., hate, sexual, violence), with higher values indicating safer CS~\cite{bonaldi2024safer}.

\noindent \textbf{LLM-as-a-Judge Evaluation.} 
We use an adaptation of JudgeLM \cite{zhu2023judgelm} tailored for CS evaluation~\cite{bonaldi2025first,zubiaga2024llm}. 
While traditional metrics capture surface and semantic alignment, JudgeLM provides a more holistic evaluation of CS quality along different dimensions, aligning with human judgements. We use the version with fine-tuned Llama-instruct-7B, following \cite{bonaldi2025first}.


\noindent \textbf{Human evaluation metrics.} Automatic metrics cannot fully account for pragmatic qualities of CS. To address this, we conduct a human evaluation focusing on dimensions directly relevant to the effectiveness and quality of CS \cite{stapleton2015assessing,bengoetxea-etal-2024-basque,bonaldi2024safer}. Participants assess each CS along the following criteria, using a Likert scale from 1 to 3 (3 being the highest score) (see Appendix \ref{app:guidelines} for full guidelines):

\noindent \textbf{Relevance}: the CS specifically addresses both the topic and the intended target of the HS.

\noindent \textbf{Correctness}: measures CS stylistic quality, including fluency and absence of offensive language.

\noindent \textbf{Factuality}: measures wheter the CS introduces additional factual information that is accurate.

\noindent \textbf{Cogency}: measures whether the CS presents logically sound and relevant arguments that effectively counter the HS.

\noindent In addition, two binary judgments are collected:

\noindent \textbf{Effectiveness}: the CS is considered persuasive and likely to influence the perspective of the HS author or audience.

\noindent \textbf{Best Response}: for each HS message, annotators select the most effective CS among multiple ones.


\begin{table*}[t]
\centering
\scriptsize
\setlength{\tabcolsep}{3pt}
\begin{tabular}{lrrrrrrrrrrrr}
\toprule
\textbf{Model} & \textbf{BLEU} & \textbf{METEOR} & \textbf{ROUGE-L} & \textbf{BERTScore} & \textbf{Distinct-1} & \textbf{Distinct-2} & \textbf{RR} & \textbf{Safety} & \textbf{Contr.} & \textbf{Ent.} & \textbf{Neutral} & \textbf{Similarity} \\
\midrule
\multicolumn{13}{c}{\textbf{No Retrieval (No-RAG)}} \\
\midrule
Llama     & .0115{\tiny$\pm$.0004} & .1638{\tiny$\pm$.0020} & \textbf{.1164}{\tiny$\pm$.0015} & .8575{\tiny$\pm$.0004} & .0196 & .1238 & .0034 & .9867{\tiny$\pm$.0010} & - & - & - & - \\
CommandR  & \textbf{.0118}{\tiny$\pm$.0004} & .1432{\tiny$\pm$.0020} & .1163{\tiny$\pm$.0016} & \textbf{.8590}{\tiny$\pm$.0004} & .0233 & .1462 & .0002 & .9884{\tiny$\pm$.0010} & - & - & - & - \\
Mistral   & .0116{\tiny$\pm$.0004} & .1435{\tiny$\pm$.0018} & .1121{\tiny$\pm$.0015} & .8575{\tiny$\pm$.0004} & .0210 & .1059 & .2668 & .9924{\tiny$\pm$.0007} & - & - & - & - \\
GPT       & .0105{\tiny$\pm$.0003} & .1494{\tiny$\pm$.0018} & .1124{\tiny$\pm$.0014} & .8580{\tiny$\pm$.0004} & .0156 & .1037 & .0018 & \textbf{.9934}{\tiny$\pm$.0006} & - & - & - & - \\
\midrule
\multicolumn{13}{c}{\textbf{BM25 Retrieval}} \\
\midrule
Llama     & .0079{\tiny$\pm$.0003} & .1639{\tiny$\pm$.0017} & .1059{\tiny$\pm$.0012} & .8491{\tiny$\pm$.0004} & .0291 & .1976 & .0088 & .9735{\tiny$\pm$.0018} & .1038{\tiny$\pm$.0040} & .2071{\tiny$\pm$.0057} & .6891{\tiny$\pm$.0063} & .5478{\tiny$\pm$.0050} \\
CommandR  & .0080{\tiny$\pm$.0002} & .1527{\tiny$\pm$.0017} & .1059{\tiny$\pm$.0012} & .8484{\tiny$\pm$.0004} & .0304 & .1963 & .0010 & .9775{\tiny$\pm$.0014} & .1351{\tiny$\pm$.0054} & .2848{\tiny$\pm$.0081} & .5801{\tiny$\pm$.0079} & .6421{\tiny$\pm$.0042} \\
Mistral   & .0079{\tiny$\pm$.0002} & .1442{\tiny$\pm$.0017} & .0986{\tiny$\pm$.0013} & .8490{\tiny$\pm$.0004} & \textbf{.0371} & \textbf{.2173} & .0004 & .9814{\tiny$\pm$.0013} & .1333{\tiny$\pm$.0060} & .2229{\tiny$\pm$.0072} & .6438{\tiny$\pm$.0076} & .6187{\tiny$\pm$.0036} \\
GPT       & .0070{\tiny$\pm$.0002} & .1580{\tiny$\pm$.0016} & .1015{\tiny$\pm$.0011} & .8507{\tiny$\pm$.0004} & .0231 & .1648 & \textbf{.0000} & .9828{\tiny$\pm$.0011} & .0567{\tiny$\pm$.0040} & .2258{\tiny$\pm$.0073} & .7175{\tiny$\pm$.0074} & .5984{\tiny$\pm$.0038} \\
\midrule
\multicolumn{13}{c}{\textbf{Sentence-BERT Retrieval}} \\
\midrule
Llama     & .0087{\tiny$\pm$.0003} & .1683{\tiny$\pm$.0018} & .1090{\tiny$\pm$.0012} & .8508{\tiny$\pm$.0004} & .0272 & .1835 & .0106 & .9705{\tiny$\pm$.0017} & .0911{\tiny$\pm$.0037} & .2135{\tiny$\pm$.0058} & .6954{\tiny$\pm$.0064} & .4698{\tiny$\pm$.0052} \\
CommandR  & .0090{\tiny$\pm$.0003} & .1614{\tiny$\pm$.0018} & .1115{\tiny$\pm$.0013} & .8504{\tiny$\pm$.0004} & .0281 & .1780 & .0054 & .9694{\tiny$\pm$.0017} & .1537{\tiny$\pm$.0055} & .2018{\tiny$\pm$.0067} & .6445{\tiny$\pm$.0070} & .5539{\tiny$\pm$.0049} \\
Mistral   & .0085{\tiny$\pm$.0002} & .1469{\tiny$\pm$.0017} & .1011{\tiny$\pm$.0013} & .8500{\tiny$\pm$.0004} & .0347 & .2009 & .0010 & .9755{\tiny$\pm$.0016} & .1386{\tiny$\pm$.0057} & .1624{\tiny$\pm$.0054} & .6990{\tiny$\pm$.0066} & .5539{\tiny$\pm$.0039} \\
GPT       & .0075{\tiny$\pm$.0002} & .1616{\tiny$\pm$.0016} & .1036{\tiny$\pm$.0011} & .8519{\tiny$\pm$.0004} & .0213 & .1531 & \textbf{.0000} & .9785{\tiny$\pm$.0013} & .0496{\tiny$\pm$.0033} & .1623{\tiny$\pm$.0057} & \textbf{.7882}{\tiny$\pm$.0060} & .5284{\tiny$\pm$.0041} \\
\midrule
\multicolumn{13}{c}{\textbf{BGE-M3 Retrieval}} \\
\midrule
Llama     & .0089{\tiny$\pm$.0003} & \textbf{.1744}{\tiny$\pm$.0019} & .1119{\tiny$\pm$.0013} & .8511{\tiny$\pm$.0004} & .0237 & .1667 & .0042 & .9720{\tiny$\pm$.0017} & .1054{\tiny$\pm$.0044} & .2115{\tiny$\pm$.0058} & .6832{\tiny$\pm$.0067} & .5864{\tiny$\pm$.0044} \\
CommandR  & .0091{\tiny$\pm$.0003} & .1597{\tiny$\pm$.0018} & .1143{\tiny$\pm$.0013} & .8523{\tiny$\pm$.0004} & .0247 & .1713 & .0002 & .9742{\tiny$\pm$.0014} & .1457{\tiny$\pm$.0055} & .2923{\tiny$\pm$.0080} & .5620{\tiny$\pm$.0078} & .6781{\tiny$\pm$.0035} \\
Mistral   & .0094{\tiny$\pm$.0003} & .1505{\tiny$\pm$.0018} & .1063{\tiny$\pm$.0013} & .8537{\tiny$\pm$.0004} & .0315 & .1939 & .0008 & .9793{\tiny$\pm$.0013} & .1329{\tiny$\pm$.0063} & .2321{\tiny$\pm$.0071} & .6350{\tiny$\pm$.0076} & .6641{\tiny$\pm$.0029} \\
GPT       & .0079{\tiny$\pm$.0002} & .1658{\tiny$\pm$.0017} & .1086{\tiny$\pm$.0012} & .8538{\tiny$\pm$.0004} & .0203 & .1533 & \textbf{.0000} & .9818{\tiny$\pm$.0011} & \textbf{.0441}{\tiny$\pm$.0036} & \textbf{.3108}{\tiny$\pm$.0083} & .6451{\tiny$\pm$.0082} & \textbf{.7055}{\tiny$\pm$.0030} \\
\bottomrule
\end{tabular}
\caption{CS automatic scores across different LLMs and retrieval strategies (averages) with 95\% bootstrap confidence intervals. Abbreviations are for: Contradiction (Contr.), Entailment (Ent.), and Repetition Rate (RR). Best results are in bold.}
\label{tab:automatic-metrics-full}
\end{table*}

\section{Results}

\textbf{Automatic metrics analysis.}
Table~\ref{tab:automatic-metrics-full} reports average results, highlighting consistent gains from incorporating RAG. 
\textbf{Content Quality:} Across all models, RAG variants maintain high semantic alignment with MT-Co references, with BERTScore remaining stable ($\approx$0.85) and comparable to No-RAG configurations. While No-RAG systems achieve slightly higher BLEU, and ROUGE-L, reflecting closer surface overlap with reference phrasing, RAG-based outputs 
systematically add factual content while preserving semantic alignment, leading to higher METEOR scores.
\textbf{Diversity:} RAG consistently improves lexical diversity and reduces repetition for every model. Distinct-1/2 scores increase with all retrieval strategies, while repetition rates drop compared to No-RAG outputs. This effect is most pronounced for Mistral, which exhibits high repetition without retrieval but becomes one of the most diverse generators with RAG. These results indicate that RAG mitigates generic responses, promoting more varied and content-rich CS. \textbf{Safety:} No-RAG configurations achieve marginally higher safety scores; still, RAG-based models retain consistently high safety. 
The slight drop likely reflects occasional noise in retrieved content rather than a systematic reduction in safety, indicating that evidence grounding does not substantially compromise non-harmfulness. \textbf{Model-level Trends:} For each LLM, RAG variants outperform their No-RAG counterparts in terms of diversity and informativeness while preserving strong semantic similarity and safety. GPT remains the most stable and safest model overall, Mistral benefits the most from retrieval, and BGE-M3 consistently provides the best trade-off across quality, diversity, and grounding. Differences are statistically significant (see Appendix \ref{app:stat_sig}).

\textbf{Hallucination analysis.}
We evaluate whether RAG reduces hallucinations by measuring the consistency between generated CS and retrieved evidence. Using \texttt{RoBERTa-large-MNLI} as a natural language inference (NLI) model, we treat retrieved evidence as the premise and CS as the hypothesis, yielding probabilities for Contradiction, Entailment, and Neutrality. Low contradiction indicates factual consistency, while entailment reflects direct factual support; neutral captures cases where CS is neither contradicted nor entailed by the evidence, as expected for explanatory responses. Additionally, we compute semantic similarity with Sentence-BERT (\texttt{all-MiniLM-L6-v2}) to measure overall grounding. 
Table~\ref{tab:automatic-metrics-full} shows that contradiction scores are consistently low across all retrievers and models ($\le 0.15$, as low as 0.04 for GPT), indicating that CS rarely contradicts evidence. Entailment scores are moderate, reflecting the expected nature of CS, which often rephrases or explains rather than directly restates evidence. Neutral scores are high indicating that much of the generated CS is neither contradicted nor directly entailed by the retrieved information.
Semantic similarity is high, especially for BGE-M3 (up to 0.71), showing strong grounding. 
Overall, these results suggest that RAG constrains CS generation to remain largely consistent with the retrieved evidence, with relatively few detected contradictions. However, NLI neutrality does not establish the absence of unsupported claims; therefore, these results should be interpreted as evidence of consistency with retrieved evidence rather than as a direct measure of hallucination.
Results are statistically significant (Appendix \ref{app:stat_sig}).

\textbf{LLM-as-a-Judge Evaluation.}
We evaluate CS quality using JudgeLM, across four dimensions: factuality, number of relevant facts, relevance to the HS, and specificity. Each evaluation is a pairwise comparison (totaling 5003) between model outputs, emphasizing informative and targeted CS over surface-level similarity (prompts are in Appendix \ref{app:prompt}).
Panel~A of Table~\ref{tab:judgelm-panels} shows wins for RAG vs. No-RAG models across retrieval strategies. RAG setups consistently outperform No-RAG counterparts, with the sole exception of CommandR + BM25, which loses slightly. BGE-M3 yields the strongest results, confirming the advantage of semantically rich retrieval, and GPT wins the majority of comparisons. 
Panel~B of Table~\ref{tab:judgelm-panels} shows all RAG setups compared to human-written MT-Co CS. All RAG systems perform strongly, with GPT + BGE-M3 being the best, followed by Mistral and Llama + BGE-M3, while CommandR lags slightly but still outperforms the baseline.
This result is particularly notable given that MT-Co is based on human-in-the-loop CS generation. The advantage of the proposed approach may partly stem from the additional factual knowledge introduced through retrieval, which can make responses more informative and better grounded. However, this comparison should be interpreted with caution, as MT-Co and our approach differ in their data collection and generation settings, and therefore existing dataset-based CS may not provide a fully comparable baseline. Further, JudgeLM may favor characteristics of LLM-generated text, such as fluency, informativeness, or stylistic consistency, potentially contributing to the observed gap. Thus, while our results show that RAG-based CS improves against MT-Co under our evaluation setting, they should not be interpreted as evidence that LLM-generated CS is inherently superior to human-written CS.

\textbf{Interpretation.}
Together, automatic metrics and JudgeLM reveal three trends: \textbf{(1)} RAG CS is less repetitive and more diverse than No-RAG while maintaining high lexical and semantic similarity to HS; \textbf{(2)} retrieved evidence is effectively incorporated, with low contradiction probabilities and high semantic similarity to the supporting content; and \textbf{(3)} RAG consistently outperforms No-RAG and MT-Co under our evaluation setting. Overall, these findings suggest that RAG improves CS quality, diversity, and factual grounding while maintaining strong alignment with HS and supporting evidence, indicating that automatic CS generation can be made sufficiently factual and non-generic to support deployment in social media settings (\textbf{RQ1}).

\begin{table}[t]
\centering
\small 
\begin{tabular}{lrrr}
\toprule
\multicolumn{4}{c}{\textbf{Panel A: JudgeLM RAG wins vs No RAG}} \\
\midrule
\textbf{Model} & \textbf{BM25} & \textbf{SentBERT} & \textbf{BGE-M3} \\
\midrule
Mistral   & 3011 & 3103 & 3557 \\
Llama     & 2576 & 2805 & 3532 \\
GPT       & \textbf{3524} & \textbf{3941} & \textbf{4223} \\
CommandR  & 2490 (lost) & 2620 & 3125 \\
\midrule
\multicolumn{4}{c}{\textbf{Panel B: JudgeLM RAG wins vs MT-Co}} \\
\midrule
\textbf{Model} & \textbf{BM25} & \textbf{SentBERT} & \textbf{BGE-M3} \\
\midrule
Mistral   & 4721 & 4831 & 4944 \\
Llama     & 4701 & 4757 & 4893 \\
GPT       & \textbf{4866} & \textbf{4948} & \textbf{4976} \\
CommandR  & 4161 & 4351 & 4653 \\
\bottomrule
\end{tabular}
\caption{Wins in pairwise comparisons for RAG vs No-RAG (Panel A) and RAG vs MT-Co (Panel B). ``Lost'' indicates the model is outperformed. Best results are in bold.}
\label{tab:judgelm-panels}
\end{table}

\section{Human Evaluation}

We conduct a human evaluation with 26 participants, voluntarily recruited, to assess the quality and effectiveness of the generated CS\footnote{Participants' age is between 18 and 50, there is a balance between genders, and their level of education spans high school diploma to PhD. Detailed statistics are in Appendix \ref{app:demo}.}. Our evaluation setup consists of 10 HS examples, extracted randomly from MT-Co keeping the target distribution, with 9 CS candidates for each HS. We select the CS from MT-Co and generated by our four LLMs without the RAG pipeline, and the corresponding CS with RAG using BGE-M3 as retriever, since it has shown the strongest performance in the automatic and LLM-based metrics. Participants rate all 90 CS candidates along four dimensions, i.e., Relevance, Factuality, Cogency, and Correctness (1-3 scale), judge whether each CS is effective (Yes/No) and select the best CS per HS. We collect a total of 2340 evaluations. Table~\ref{tab:metric_scores} reports average scores for quality metrics.






\noindent \textbf{Metric Scores.} RAG-based methods consistently outperform their No-RAG counterparts. CommandR RAG achieves the highest Relevance (2.75), while Llama RAG leads in Factuality (2.50). GPT RAG achieves the highest Cogency (2.55), ties for the top Correctness (2.75), while MT-Co performs worst across all dimensions.


\begin{table}[t!]
\centering
\scriptsize
\setlength{\tabcolsep}{4pt}
\begin{tabular}{lrrrrrr}
\toprule
\textbf{Method} & \textbf{Rel.} & \textbf{Fact.} & \textbf{Cog.} & \textbf{Corr.} & \textbf{Eff.} & \textbf{Best} \\
\midrule
MT-Co & 2.07{\tiny$\pm$0.74} & 1.38{\tiny$\pm$0.67} & 1.67{\tiny$\pm$0.73} & 2.28{\tiny$\pm$0.82} & 48  & 7 \\
Llama NoRAG & 2.72{\tiny$\pm$0.59} & 2.14{\tiny$\pm$0.78} & 2.49{\tiny$\pm$0.65} & 2.77{\tiny$\pm$0.56} & 171 & 33 \\
Llama RAG & 2.66{\tiny$\pm$0.64} & \textbf{2.50}{\tiny$\pm$0.66} & 2.53{\tiny$\pm$0.69} & 2.75{\tiny$\pm$0.57} & \textbf{197} & 67 \\
Com.R NoRAG & 2.63{\tiny$\pm$0.62} & 1.85{\tiny$\pm$0.75} & 2.25{\tiny$\pm$0.71} & 2.75{\tiny$\pm$0.58} & 120 & 7 \\
Com.R RAG & \textbf{2.75}{\tiny$\pm$0.55} & 2.25{\tiny$\pm$0.74} & 2.48{\tiny$\pm$0.63} & 2.74{\tiny$\pm$0.57} & 172 & 26 \\
Mistral NoRAG & 2.58{\tiny$\pm$0.67} & 1.77{\tiny$\pm$0.77} & 2.11{\tiny$\pm$0.75} & 2.69{\tiny$\pm$0.63} & 109 & 11 \\
Mistral RAG & 2.60{\tiny$\pm$0.66} & 2.16{\tiny$\pm$0.70} & 2.30{\tiny$\pm$0.72} & 2.69{\tiny$\pm$0.64} & 156 & 24 \\
GPT NoRAG & 2.60{\tiny$\pm$0.62} & 1.89{\tiny$\pm$0.76} & 2.32{\tiny$\pm$0.69} & 2.72{\tiny$\pm$0.62} & 148 & 14 \\
GPT RAG & 2.68{\tiny$\pm$0.62} & 2.41{\tiny$\pm$0.73} & \textbf{2.55}{\tiny$\pm$0.67} & \textbf{2.75}{\tiny$\pm$0.56} & 191 & \textbf{71} \\
\bottomrule
\end{tabular}
\caption{Average scores for Relevance, Factuality, Cogency, and Correctness, with Effectiveness and Best CS votes.}
\label{tab:metric_scores}
\end{table}
\textbf{Best CS.} RAG-based methods are chosen more frequently as best CS, implying that they are clearly preferred by humans compared to their No-RAG counterparts, with GPT RAG receiving the highest number of votes (71). 
\textbf{Effectiveness.} Out of the 2340 evaluations collected, 1312 are marked as effective, corresponding to 56\% of the cases. More specifically, Llama RAG is considered effective most frequently (with 197 votes), followed closely by GPT RAG (191) and CommandR RAG (172). The baseline method MT-Co receives the fewest effectiveness votes (48). 
This indicates that RAG-based methods generally produce more effective CS than their No-RAG counterparts and the baseline, suggesting that both model choice and augmentation strategy strongly influence performance.


\begin{table}[t]
\centering
\footnotesize
\begin{tabular*}{\columnwidth}{@{\extracolsep{\fill}}lcc}
\toprule
\multicolumn{3}{l}{\textbf{Panel A: Average ratings by dimension and RAG status}} \\
\midrule
\textbf{Dimension / RAG} & \textbf{In Mean} & \textbf{Out Mean}  \\
\midrule
Cogency (No RAG)     & 2.10 $\pm$ 0.77 & 2.32 $\pm$ 0.69 \\
Cogency (RAG)        & \underline{2.43} $\pm$ 0.71 & \textbf{2.54} $\pm$ 0.57 \\
Correctness (No RAG) & 2.58 $\pm$ 0.72 & 2.77 $\pm$ 0.54 \\
Correctness (RAG)    & \underline{2.71} $\pm$ 0.62 & \textbf{2.80} $\pm$ 0.49 \\
Factuality (No RAG)  & 1.71 $\pm$ 0.79 & 2.01 $\pm$ 0.73 \\
Factuality (RAG)     & \underline{2.25} $\pm$ 0.74 & \textbf{2.50} $\pm$ 0.63 \\
Relevance (No RAG)   & 2.44 $\pm$ 0.67 & 2.71 $\pm$ 0.56 \\
Relevance (RAG)      & \underline{2.61} $\pm$ 0.67 & \textbf{2.80} $\pm$ 0.46 \\
\midrule
\multicolumn{3}{l}{\textbf{Panel B: Mean Effectiveness Rates by RAG status}} \\
\midrule
\textbf{RAG Status} & \textbf{In-group} & \textbf{Out-group}  \\
\midrule
No RAG  & 0.43 $\pm$ 0.19 & 0.53 $\pm$ 0.29 \\
RAG     & \textbf{0.68} $\pm$ 0.17 & \textbf{0.71} $\pm$ 0.28 \\
\bottomrule
\end{tabular*}

\caption{Quality metrics across RAG status and in/out-group participants (Panel A) and effectiveness rates (Panel B). Best results are in \textbf{bold} and second best \underline{underlined}.}
\label{tab:cs_combined_results_updated}
\end{table}

To address LLM-as-a-Judge reliability, we analyze concordance between JudgeLM pairwise outcomes and human quality scores using Spearman correlation. LLM-human $\rho$ is directionally positive (up to 0.40), while inter-metric agreement among human dimensions is strong and significant ($\rho$: 0.545--0.983), confirming the internal consistency of the evaluation framework. Overall, RAG-based methods consistently outperform No-RAG counterparts across all quality metrics, receive higher effectiveness ratings, and are more often chosen as best. Both human evaluation and JudgeLM consistently identify RAG as the strongest system and MT-Co as the weakest across all dimensions, further supporting the reliability of the automatic judge. We additionally assessed the relationship between automatic and human evaluation through Spearman correlations at the HS$\times$CS-instance level. Several moderate associations were observed, but none remained significant after Holm correction, suggesting that automatic and human evaluation provide complementary perspectives on CS quality, while the relatively small sample size may limit statistical power (full results in Appendix \ref{appendix:full_stats}).



\section{Social Media Analysis and Discussion}

Automatic CS is intended for deployment in social media environments, where it is addressed not only to the HS author but also to bystanders and members of targeted communities. Aggregate quality metrics, however, provide limited insight into how CS is viewed by different audiences. We therefore analyze how CS is perceived by different groups and discuss implications for platform deployment.

\noindent \textbf{In- and Out-Group Evaluation.} To answer RQ2, we classify annotators as in-group if they self-identify with at least one group targeted by HS in our evaluation set\footnote{In-group identities include women (13), LGBT+ (5), migrants (4), Muslims (2), and people of color (1); some participants reported multiple identities, while two preferred not to disclose.}. Across Relevance, Factuality, Cogency, and Correctness, in-group annotators assign lower ratings than out-group annotators for both RAG and No-RAG systems (Panel~A of Table~\ref{tab:cs_combined_results_updated}). Mixed-effects analyses (Appendix~F) indicate that this difference is not explained by group identity alone after accounting for CS strategy and message-level variability, with significant interactions suggesting that in-group annotators are more selective across generation strategies. 
Effectiveness judgments show a similar pattern. Panel~B reports mean effectiveness rates (average of binary effectiveness judgments, aggregated at the HS level, then averaged across HS within each group). RAG models are rated as more effective by both groups, with a larger absolute gain for in-group evaluators (0.43 to 0.68). 
Although these results should be interpreted cautiously given the limited and uneven sample, they suggest that evaluations based only on aggregate or predominantly out-group judgments may overlook differences in how CS is received by audiences connected to the targeted communities.
In-group annotators also exhibit lower entropy in their selection of preferred generation strategies (0.93 vs.\ 1.56), with 62\% of their top choices concentrated on a smaller subset of strategies, compared with 40\% for out-group annotators (Appendix~F, Table \ref{tab:cs_combined_full}). This suggests that community membership may be associated not only with stricter evaluations, but also with more concentrated preferences regarding what constitutes an acceptable response. For platforms, this motivates evaluating automated CS across different audiences rather than relying exclusively on aggregate judgments. It also suggests that a single generation strategy may not be equally appropriate across communities.
The KB composition poses an additional consideration for platform scale deployment. UN and EU sources provide uneven coverage across target groups, so the amount and type of institutional evidence may vary across communities. Moreover, their formal, policy-oriented language may be less aligned with the conversational norms of social media than news or community generated sources, potentially reducing perceived relevance or authenticity among targets and bystanders. Institutional grounding therefore offers greater authority and traceability, but may trade off some social media fit. While our results support its effectiveness for grounded CS, they do not establish that institutional knowledge is always optimal for real world deployment.

\noindent \textbf{Implications and limitations.} Our findings suggest three considerations for deploying automated CS on social media. First, evaluation should include members of the communities targeted by HS, as aggregate judgments may mask differences in perceived quality and effectiveness. Second, factual grounding can improve the perceived effectiveness of CS, but the remaining gap between in-group and out-group evaluations indicates that grounding alone does not guarantee community acceptance. Third, automated CS should be considered a complement to, rather than a replacement for, expert- and community-informed approaches, particularly for high stakes cases. Importantly, our evaluation presents CS as isolated messages rather than interactions embedded in comment threads, reply chains, or recommendation environments. It therefore cannot establish how users or bystanders would respond to CS online, nor whether generated responses would affect subsequent engagement or exposure to HS. Larger evaluations with more balanced representation across target groups and studies conducted in realistic social media settings are needed to assess these effects.

\section{Comparison with RAG-based CS Systems}

\begin{table}[t]
\centering
\scriptsize
\begin{tabular}{lrrrr}
\toprule
\textbf{Competitor} & \textbf{Target} & \textbf{Total} & \textbf{Wins (ours)} & \textbf{\% Wins} \\
\midrule
\multirow{8}{*}{vs Wilk et al. (2025)} 
 & DISABLED & 36  & 25  & 69.4 \\
 & JEWS     & 204 & 102 & 50.0 \\
 & LGBT+    & 164 & 105 & 64.0 \\
 & MIGRANTS & 310 & 176 & 56.8 \\
 & MUSLIMS  & 518 & 286 & 55.2 \\
 & POC      & 109 & 58  & 53.2 \\
 & WOMEN    & 232 & 142 & 61.2 \\
 & Other    & 124 & 68  & 54.8 \\
 & \textbf{Total} & 1697 & 962 & \textbf{57.0} \\
\midrule
\multirow{5}{*}{vs Russo (2025)} 
 & JEWS     & 23  & 22  & 95.7 \\
 & LGBT+    & 19  & 17  & 89.5 \\
 & MIGRANTS & 30  & 30  & 100.0 \\
 & POC      & 15  & 11  & 73.3 \\
 & WOMEN    & 37  & 34  & 91.9 \\
 & \textbf{Total} & 124 & 114 & \textbf{93.0} \\
\midrule
\multirow{5}{*}{vs Bonaldi et al. (2025)} 
 & JEWS     & 83  & 82  & 98.8 \\
 & LGBT+    & 85  & 83  & 97.6 \\
 & MIGRANTS & 114 & 114 & 100.0 \\
 & POC      & 73  & 73  & 100.0 \\
 & WOMEN    & 141 & 141 & 100.0 \\
 & \textbf{Total} & 496 & 493 & \textbf{99.4} \\
\bottomrule
\end{tabular}
\caption{JudgeLM pairwise wins per target for our RAG-based CS against three competitors.}
\label{tab:per-target-comparison}
\end{table}

To contextualize our results, we compare our RAG-based CS with publicly available samples from three recent studies. All CS are aligned with MT-Co HS, and we retain only the English ones. \citet{wilk2025fact} (Baseline 1) released 1,697 GPT-4o RAG-generated CS, using \textit{Newsroom} as the primary knowledge source and the Tavily web-search API to retrieve more recent documents. \citet{russo2025trenteam} (Baseline 2) contributed 124 LLaMA-EUS-8B RAG-generated CS, retrieving evidence from \textit{Newsroom} and \textit{WikiText-103}. \citet{bonaldi2025first} (Baseline 3) provided 496 triplets of HS, CS, and external knowledge, using a human-expert hybrid RAG system. For fair comparison, we reproduce settings closest to the original ones, i.e., GPT-4o-mini + BGE-M3 for Baselines 1 and 3, and Llama-8B + BGE-M3 for Baseline 2, isolating the effects of retrieval and LLM design.
Pairwise performance is evaluated with JudgeLM (prompt in Appendix~\ref{app:prompt}), which rates helpfulness, relevance, accuracy, and level of detail, while also considering conciseness and social media suitability. Table~\ref{tab:per-target-comparison} shows that our RAG pipeline consistently outperforms all baselines. GPT + BGE-M3 wins 962 out of 1,697 comparisons against Baseline 1 and 493 out of 496 against Baseline 3, while Llama + BGE-M3 wins 114 out of 124 against Baseline 2.
Table~\ref{tab:per-target-comparison} shows that the advantage is generally consistent across HS targets. Against Baseline 1, gains are observed across all categories, ranging from 50\% for \textit{JEWS} to 69.4\% for \textit{DISABLED}. Against Baseline 2, our system wins 93\% of comparisons, while against Baseline 3 it wins 99.4\%. These results indicate that the observed advantage is not restricted to a particular target group, although differences in baseline construction and sample size require cautious interpretation.

The advantage over prior RAG systems may partly reflect differences in the evidence used for grounding. News and web sources provide broad and potentially current information, while our KB focuses on authoritative institutional sources from international and regional human rights organizations, offering more directly relevant and traceable evidence for CS. At the same time, differences in retrieval, generation, and evaluation settings prevent us from attributing the gains to knowledge source alone. The results therefore suggest that the combination of authoritative knowledge, retrieval, and generation design contributes to stronger CS, rather than demonstrating that any single knowledge source is universally superior.

\begin{table*}[t]
\centering
\footnotesize
\begin{tabular}{llrrrrrrr}
\toprule
\textbf{Group} & \textbf{Setting} & \textbf{Specificity} & \textbf{Evidence} & \textbf{Conciseness} & \textbf{Persuasiveness} & \textbf{Grammaticality} & \textbf{CS Tokens} & \textbf{Input Tokens} \\
\midrule

\multirow{6}{*}{$k$}
& 0 & 4.51 $\pm$ 0.54 & 0.00 $\pm$ 0.00 & 4.43 $\pm$ 0.45 & 4.73 $\pm$ 0.31 & 5.00 $\pm$ 0.00 & 34.88 & -- \\
& 1 & 4.73 $\pm$ 0.44 & 4.34 $\pm$ 1.12 & 3.79 $\pm$ 0.41 & 4.76 $\pm$ 0.31 & 5.00 $\pm$ 0.00 & 80.70 & -- \\
& 2 & 4.74 $\pm$ 0.48 & 4.31 $\pm$ 1.09 & 3.74 $\pm$ 0.44 & 4.73 $\pm$ 0.32 & 5.00 $\pm$ 0.00 & 84.08 & -- \\
& 3 & 4.79 $\pm$ 0.47 & 4.42 $\pm$ 0.95 & 3.74 $\pm$ 0.46 & 4.75 $\pm$ 0.33 & 5.00 $\pm$ 0.00 & 89.29 & -- \\
& 4 & 4.79 $\pm$ 0.44 & 4.48 $\pm$ 0.81 & 3.68 $\pm$ 0.50 & 4.79 $\pm$ 0.30 & 4.98 $\pm$ 0.11 & 92.16 & -- \\
& 5 & 4.84 $\pm$ 0.44 & 4.64 $\pm$ 0.62 & 3.70 $\pm$ 0.47 & 4.84 $\pm$ 0.31 & 5.00 $\pm$ 0.00 & 95.38 & -- \\

\midrule

\multirow{2}{*}{Summ.}
& Full  & 4.65 $\pm$ 0.46 & 4.04 $\pm$ 1.42 & 3.49 $\pm$ 0.50 & 4.67 $\pm$ 0.47 & 3.18 $\pm$ 0.31 & 174.2 & 184 \\
& Sum.  & \textbf{4.79} $\pm$ 0.47 & \textbf{4.42} $\pm$ 0.95 & \textbf{3.74} $\pm$ 0.46 & \textbf{4.76} $\pm$ 0.33 & 5.00 $\pm$ 0.00 & 53.80 & 131.5 \\

\midrule

\multirow{3}{*}{Data}
& EU      & 4.51 $\pm$ 0.57 & 3.39 $\pm$ 1.34 & \textbf{3.99} $\pm$ 0.34 & 4.68 $\pm$ 0.24 & 5.00 $\pm$ 0.00  & 51.55 & -- \\
& UN      & 4.54 $\pm$ 0.54 & 3.66 $\pm$ 1.30 & 3.91 $\pm$ 0.31 & 4.69 $\pm$ 0.33 & 5.00 $\pm$ 0.00 & 53.14 & -- \\
& EU+UN   & \textbf{4.79} $\pm$ 0.47 & \textbf{4.42} $\pm$ 0.95 & 3.74 $\pm$ 0.46 & \textbf{4.75} $\pm$ 0.33 & 5.00 $\pm$ 0.00 & 53.80 & -- \\

\bottomrule
\end{tabular}

\caption{Ablation results across n. of retrieved paragraphs ($k$), summarization settings, and KB sources. Best results are in bold.}
\label{tab:combined_ablation_compact}
\end{table*}

\section{Ablation Study}

We conduct an ablation study to assess the impact of RAG pipeline components: the number of retrieved evidence paragraphs ($k$), the summarization step, and KB composition. We construct a stratified random sample of 10 HS per target category using a fixed random seed, then shuffle the subset to avoid ordering effects, yielding 80 HS messages.

\noindent \textbf{Effect of number of retrieved paragraphs.} For each HS, CS is generated using $k \in \{0,1,2,3,4,5\}$ paragraphs retrieved by BGE-M3 ($k=0$ corresponds to no evidence, No-RAG), using GPT-4o-mini and a fixed prompt template as in the main experiment, obtaining a total of 480 generations.
We evaluate CS quality on specificity, evidence utilization, conciseness, persuasiveness, and grammar, computed in an LLM-as-a-Judge setting using two models (GPT-4o-mini and Mistral), averaging the results. The top of Table~\ref{tab:combined_ablation_compact} reports scores mean $\pm$ standard deviation for different values of $k$. Across most metrics, performance improves from $k=1$ to $k=3$ and plateaus thereafter. Grammar scores are consistently perfect (5.0) except for $k=4$ (4.98). Conciseness slightly decreases beyond $k=3$, reflecting longer, more verbose outputs without substantial gains in CS quality. Overall, using three evidence paragraphs ($k=3$) achieves the optimal balance, maximizing evidence grounding and persuasiveness while maintaining concise, high-quality text, and this setting is adopted in our main experiments.

\noindent \textbf{Effect of summarization.} We use the same 80 hateful messages from the previous ablation study and we generate CS using our best LLM (GPT-4o-mini) and retriever (BGE-M3), by using the full retrieved paragraphs, and the summarized ones, for a total of 160 generated CS. For evaluation, the same quality scores are computed in the same LLM-as-a-Judge setting. In addition, we measure token counts of the generated CS and the input paragraphs (using \texttt{tiktoken} and GPT-4o-mini's encoder, i.e. \texttt{o200k\_base}), and assess their grammatical correctness. The middle of Table~\ref{tab:combined_ablation_compact} shows that summarized paragraphs produce significantly fewer CS tokens (53.8 vs 174.2), improving efficiency and making messages easier to process, which is reflected in Conciseness which rises from 3.49 to 3.74, showing that summarization cuts unnecessary words while retaining essential content. The same pattern holds also for input paragraphs tokens, where summarized ones have a lower token count. Grammar scores also increase from 3.18 to 5. 
Furthermore, also Specificity, Persuasiveness and, especially, Evidence utilization improve. These results demonstrate that the summarization step is essential for large-scale CS generation, especially when using multiple LLMs and retrievers. In our case, with four LLMs and three retrievers, we perform 12 CS generations per message for 5,003 messages (60,036 total CS); by reducing prompt length, improving grammar, and improving quality, summarized paragraphs enhance efficiency, reduce API costs, and produce professional and readable CS.

\noindent \textbf{Effect of KB components.} We evaluate the contribution of each knowledge source by comparing three retrieval conditions: EU-only (266 documents), UN-only (32,526 documents), and the combined KB (EU+UN). The bottom of Table~\ref{tab:combined_ablation_compact} reveals a complementarity between sources. EU-only achieves competitive specificity (4.51) and persuasiveness (4.68) thanks to its small corpus of dense, targeted institutional reports, but shows the lowest evidence utilization (3.39), indicating that retrieved content, while topically precise, is harder to ground. 
UN-only improves evidence use (3.66) through broader document coverage, at a marginal cost to conciseness. Finally, the combined KB achieves the highest evidence utilization (4.42) and persuasiveness (4.75) across all conditions, while maintaining strong specificity (4.79). 
EU depth and UN breadth together enable retrieval that is both topically targeted and factually grounded, which is the core objective of our RAG pipeline (additional complementarity results are in Appendix \ref{app:ablation}. Based on these results, we adopt the combined KB in all main experiments.

\section{Conclusion}

We propose a RAG framework for automatic CS generation, designed to produce factually grounded, concise, and socially responsible responses to online HS. Regarding \textbf{RQ1}, our results show that grounding generation in authoritative knowledge can make automatic CS more factual and less generic. To this end, we construct the first large, curated KB from international and regional human rights institutions, enabling reliable and evidence-based generation. Regarding \textbf{RQ2}, our human evaluation shows that CS is perceived differently by in-group and out-group annotators, with RAG substantially narrowing the gap in perceived effectiveness. Overall, our extensive automatic, LLM-based, and human evaluations, together with comparisons with recent state-of-the-art systems, support knowledge-grounded generation as a promising scalable complement to existing approaches for addressing online HS.


\section{Limitations}
First, our paragraph-level retrieval may fragment contextual information, although we mitigate this by using overlapping chunks. Restricting the retrieved context to the top-$k=3$ paragraphs also improves efficiency and conciseness but may affect the factual richness and diversity of generated CS. Second, retrieval quality depends on the coverage and relevance of the KB; gaps or biases in external sources may propagate to generated responses, despite relying on recognized authoritative sources. Third, LLMs may still produce partially hallucinated or biased outputs when retrieved evidence is noisy or ambiguous, although our factuality analysis shows very low hallucination rates. Limiting counter-speech to two sentences improves realism and readability for social media but may reduce nuance. Finally, effectiveness may vary across cultural contexts, and outputs can still contain inaccuracies despite factuality checks and human oversight; the system is intended for research use only.

\bibliography{aaai25}

\section{AAAI ICWSM Paper Checklist}

\begin{enumerate}

\item For most authors...
\begin{enumerate}
    \item  Would answering this research question advance science without violating social contracts, such as violating privacy norms, perpetuating unfair profiling, exacerbating the socio-economic divide, or implying disrespect to societies or cultures?
    \answerYes{Yes}
  \item Do your main claims in the abstract and introduction accurately reflect the paper's contributions and scope?
    \answerYes{Yes}
   \item Do you clarify how the proposed methodological approach is appropriate for the claims made? 
    \answerYes{Yes, see Section ``Knowledge Base Construction'' and ``Pipeline''.}
   \item Do you clarify what are possible artifacts in the data used, given population-specific distributions?
    \answerYes{Yes, see Section ``Experimental Setup''.}
  \item Did you describe the limitations of your work?
    \answerYes{Yes, see Section ``Limitations''.}
  \item Did you discuss any potential negative societal impacts of your work?
    \answerYes{Yes, see ``Limitations''.}
      \item Did you discuss any potential misuse of your work?
    \answerYes{Yes, see ``Limitations''.}
    \item Did you describe steps taken to prevent or mitigate potential negative outcomes of the research, such as data and model documentation, data anonymization, responsible release, access control, and the reproducibility of findings?
    \answerYes{Yes.}
  \item Have you read the ethics review guidelines and ensured that your paper conforms to them?
    \answerYes{Yes.}
\end{enumerate}

\item Additionally, if your study involves hypotheses testing...
\begin{enumerate}
  \item Did you clearly state the assumptions underlying all theoretical results?
    \answerNA{N/A}
  \item Have you provided justifications for all theoretical results?
    \answerNA{N/A}
  \item Did you discuss competing hypotheses or theories that might challenge or complement your theoretical results?
    \answerNA{N/A}
  \item Have you considered alternative mechanisms or explanations that might account for the same outcomes observed in your study?
    \answerNA{N/A} 
  \item Did you address potential biases or limitations in your theoretical framework?
    \answerNA{N/A}
  \item Have you related your theoretical results to the existing literature in social science?
    \answerNA{N/A}
  \item Did you discuss the implications of your theoretical results for policy, practice, or further research in the social science domain?
    \answerNA{N/A}
\end{enumerate}

\item Additionally, if you are including theoretical proofs...
\begin{enumerate}
  \item Did you state the full set of assumptions of all theoretical results?
    \answerNA{N/A}
	\item Did you include complete proofs of all theoretical results?
    \answerNA{N/A}
\end{enumerate}

\item Additionally, if you ran machine learning experiments...
\begin{enumerate}
  \item Did you include the code, data, and instructions needed to reproduce the main experimental results (either in the supplemental material or as a URL)?
    \answerYes{Yes, in an anonymous GitHub repository, the link is in the ``Introduction''. We release a sample of the data and the full datasets together with the knowledge base upon paper acceptance, as well as the code.}
  \item Did you specify all the training details (e.g., data splits, hyperparameters, how they were chosen)?
    \answerYes{Yes, see Section ``Pipeline'' and Appendix B.}
     \item Did you report error bars (e.g., with respect to the random seed after running experiments multiple times)?
    \answerYes{In the Tables with numerical results.}
	\item Did you include the total amount of compute and the type of resources used (e.g., type of GPUs, internal cluster, or cloud provider)?
    \answerYes{Yes, see Appendix B.}
     \item Do you justify how the proposed evaluation is sufficient and appropriate to the claims made? 
    \answerYes{Yes. See Sections 7, 8, 9, 10.}
     \item Do you discuss what is ``the cost`` of misclassification and fault (in)tolerance?
    \answerNA{N/A}
  
\end{enumerate}

\item Additionally, if you are using existing assets (e.g., code, data, models) or curating/releasing new assets, \textbf{without compromising anonymity}...
\begin{enumerate}
  \item If your work uses existing assets, did you cite the creators?
    \answerYes{Yes.}
  \item Did you mention the license of the assets?
    \answerYes{Yes.}
  \item Did you include any new assets in the supplemental material or as a URL?
    \answerYes{Yes. See URL in the ``Introduction''.}
  \item Did you discuss whether and how consent was obtained from people whose data you're using/curating?
    \answerNA{N/A. They are publicly available.}
  \item Did you discuss whether the data you are using/curating contains personally identifiable information or offensive content?
    \answerNA{N/A. The data do not contain personal information.}
\item If you are curating or releasing new datasets, did you discuss how you intend to make your datasets FAIR (see \citet{fair})?
\answerYes{Yes. We release a sample of the data with an anonymous GitHub repository publicly available, in the common CSV format and with a description of the data, see ``Introduction''. We will release all the CS generations with their retrieved evidence and the knowledge base upon paper acceptance.}
\item If you are curating or releasing new datasets, did you create a Datasheet for the Dataset (see \citet{gebru2021datasheets})? 
\answerYes{Yes, see the Datasheet in the anonymous repository.}
\end{enumerate}

\item Additionally, if you used crowdsourcing or conducted research with human subjects, \textbf{without compromising anonymity}...
\begin{enumerate}
  \item Did you include the full text of instructions given to participants and screenshots?
    \answerYes{Yes, the guidelines are in Appendix E.}
  \item Did you describe any potential participant risks, with mentions of Institutional Review Board (IRB) approvals?
    \answerYes{Yes, we describe the ethical implications of using human subjects, even if we didn't need an IRB.}
  \item Did you include the estimated hourly wage paid to participants and the total amount spent on participant compensation?
    \answerNA{N/A. Participation was voluntary.}
   \item Did you discuss how data is stored, shared, and deidentified?
   \answerYes{Yes. All human evaluations are anonymous.}
\end{enumerate}

\end{enumerate}

\appendix

\section{Keywords Used for the Knowledge Base Queries}\label{app:keywords}

Keywords were used as search queries in the three knowledge sources (UN Digital Library, EUR-Lex, and the European Union Agency for Fundamental Rights), rather than being matched through an exact or substring matching procedure implemented in our code. Each target keyword was submitted as a separate query to the corresponding repository, with multi word keywords preserved as queries (e.g., ``LGBT+ rights''). We deliberately adopted this query based strategy to favor high recall and avoid missing potentially relevant documents associated with related terminology, such as ``gay'' or ``AIDS''. The retrieved documents were subsequently deduplicated, so overlapping results across queries did not result in duplicate documents in the final knowledge base. Thus, queries such as “LGBT+” and “LGBT+ rights” were treated as distinct queries and could retrieve overlapping sets of documents, while deduplication ensured that each document was retained only once.
The following keywords were used to retrieve documents from the United Nations Digital Library, the FRA and EUR-Lex portals, covering the period 2000–2025 and targeting multiple groups and thematic areas relevant to our study:

\paragraph{Target Groups:}

\begin{itemize}
    \item \textbf{People of Color}: People of color, Racism, Anti-Black racism, Systemic racism, Racial inequality, Racial equality, Racial profiling, White privilege, Black Lives Matter, Colonialism, Racial discrimination, discrimination against black people, blacks, race, black women, black people, blacks hate speech, african descent, ethnic minorities, ethnic inequalities, minority
    \item \textbf{LGBT+}: LGBT+ rights, Homophobia, Transphobia, Biphobia, Gender identity, Conversion therapy, Same-sex marriage, Stonewall riots, LGBT+, LGBTQIA+, Gay, Lesbian, Transgender, Non-binary, LGBT+ hate speech, discrimination against LGBT+ people, gay rights movement, LGBT discrimination, LGBT+ hate crimes, sexual orientation, HIV/AIDS \& gay/lesbian
    \item \textbf{Disabled}: Disability rights, Accessibility, Social model of disability, disability, disabled, down syndrome, autism, mental disability, physical disability, neurodiversity, ableism, inclusive design, Discrimination against people with disabilities
    \item \textbf{Muslims}: Islamophobia, Discrimination against Muslims, Anti-Muslim hate crimes, Muslim communities, Religious discrimination, islam, muslim, muslim hate speech, religion, discrimination against muslims, muslim communities
    \item \textbf{Jews}: Antisemitism, Jewish identity, Anti-Jewish violence, Nazi propaganda, jews, jews hate speech, antisemitism hate speech, judaism, hebrews, jews hate crimes, jewish history, jewish diaspora, zionism, zionist movement, holocaust, israel, holocaust denial
    \item \textbf{Women}: Sexism, Misogyny, Feminism, Gender inequality, Women's rights, Me Too movement, Gender-based violence, women, women hate speech, feminism, violence against women, gender inequality, glass ceiling, discrimination against women
    \item \textbf{Migrants}: Xenophobia, Anti-immigration, Refugee crisis, Asylum seekers, Undocumented immigrants, Immigration law, refugee, migrants, immigrants, immigration, migration, immigration hate speech, migrants rights, illegal aliens, immigration and crime, immigration and unemployment, discrimination against migrants, aliens
\end{itemize}

\paragraph{Thematic Categories:}
\begin{itemize}
    \item \textbf{Hate Speech}: hate speech, hate speech laws, hate crime, hate crime legislation, hate speech regulation, hate speech prevention, online hate speech, online harassment, cyberbullying, censorship, freedom of expression, speech ethics, disinformation, radicalization, extremism, online moderation
    \item \textbf{Human Rights and Law}: human rights, human rights treaties, universal declaration of human rights, international human rights law, civil rights, social justice, equality before the law, international court of justice, refugee rights, minority rights, gender equality law, european charter of human rights
\end{itemize}

\begin{table}[t]
\centering
\small
\setlength{\tabcolsep}{3pt}
\begin{tabular}{@{}p{0.16\linewidth}p{0.68\linewidth}r@{}}
\toprule
\textbf{Target} & \textbf{Representative document} & \textbf{Year} \\
\midrule
Jews &
History under attack: holocaust denial and distortion on social media \textit{(UN Digital Library)} & 2022 \\[2pt]

LGBT+ &
EU LGBTIQ Survey III, LGBTIQ Equality at a Crossroads - Progress and Challenges \textit{(FRA)} & 2024 \\[2pt]

Migrants &
Directive 2013/33/EU, laying down standards for the reception of applicants for international protection \textit{(EUR-Lex)} & 2013 \\[2pt]

Muslims &
Being Muslim in the EU - Experiences of Muslims, EU Survey on Immigrants and Descendants of Immigrants \textit{(FRA)} & 2024 \\[2pt]

Women &
Sex-based violence against women and girls: new frontiers and emerging issues, report of the Special Rapporteur on Violence against Women and Girls, Its Causes and Consequences \textit{(UN Digital Library)} & 2025 \\[2pt]

POC &
Addressing Racism in Policing \textit{(FRA)} & 2024 \\[2pt]

Disabled &
CRPD/C/GC/8, General Comment No.\,8 on the right of persons with disabilities to work and employment \textit{(UN Digital Library)} & 2022 \\
\bottomrule
\end{tabular}
\caption{Representative documents from the KB, illustrating source and content diversity across target groups.}
\label{tab:kb_examples}
\end{table}

\subsection{Examples of Retrieved Evidence}
To illustrate the nature of the knowledge base's content, Table \ref{tab:appendix_source_breakdown} lists representative documents for a subset of target groups, spanning all three institutional sources. These range from binding legal instruments (e.g., EU directives) to survey-based empirical reports and authoritative interpretive guidance (UN treaty general comments and FRA), reflecting the mixed factual and policy-oriented character of the KB described in Section 3.

\section{Prompts used}
\label{app:prompt}

In this section, we list all the prompts used in our experiments. 

\definecolor{promptblue}{RGB}{0,90,150}
\definecolor{promptbg}{RGB}{247,249,252}

\tcbset{
  promptbox/.style={
    enhanced,
    breakable,                
    colback=white, 
    colframe=black, 
    coltitle=black,
    fonttitle=\bfseries,
    title={Prompt for keyword retrieval},
    boxrule=0.5pt,
    arc=3pt,
    top=5pt,
    bottom=5pt,
    left=7pt,
    right=7pt,
    attach boxed title to top left={xshift=4pt,yshift=-2mm},
    boxed title style={
      colback=white, 
      colframe=black, 
      arc=2pt,
      boxrule=0.5pt
    }
  }
}

To expand the set of keywords for the knowledge base creation, we prompt \texttt{GPT-5} with the following:

\begin{tcolorbox}[promptbox]
\itshape
We are conducting a thematic search across major institutional databases — the UN Digital Library, EUR-Lex (EU law), and the EU Agency for Fundamental Rights (FRA). Our goal is to identify documents, resolutions, laws, and reports related to discrimination, rights, and inclusion of specific social groups that are the targets of MultiTarget CONAN dataset: \textbf{women, people of color, persons with disabilities, migrants, Muslims, Jews, LGBT persons, other marginalized or minority groups}.

Please suggest related and institutionally accurate keywords or controlled vocabulary terms used in UN, EU, and FRA contexts for each target group. For each group, include: Formal institutional terminology (used in UN or EU documents); Common synonyms and variants (used in academic or policy discourse); Relevant policy or legal frameworks (e.g., conventions, charters, declarations); Suggested Boolean search strings for use in UN Digital Library, EUR-Lex, and FRA databases. Prioritize terminology that aligns with international human rights language and European policy frameworks (e.g., CERD, CEDAW, CRPD, FRA reports).
\end{tcolorbox}

For Step 2 of our pipeline (document summarization), for each of the 3 paragraphs, we prompt the LLMs with the following:

\begin{tcolorbox}[promptbox, title={Summarization prompt}]
\itshape
Summarize this paragraph \texttt{{paragraph\_text}} into exactly 2 sentences, without truncating the last sentence:

Summary:
\end{tcolorbox}

For Step 3 of our pipeline (CS generation), we prompt all LLMs with the following:

\begin{tcolorbox}[promptbox, title={CS generation W/OUT RAG}]
\itshape
Hateful message: \texttt{{hateful\_message}}
Generate a counter-speech that is concise, respectful, persuasive, self-contained, and suitable for social media.
Keep it short (maximum 2 sentences). Do NOT truncate the last sentence.

Counter-speech:
\end{tcolorbox}

\begin{tcolorbox}[promptbox, title={CS generation WITH RAG}]
\itshape
Here are three evidence summaries you MUST use to inform your response: \texttt{{context}}.

Task: Respond to the hateful message below by writing a short counter-speech that is respectful, persuasive, self-contained, and suitable for social media (maximum 2 sentences). Ground your response in the evidence above and do NOT truncate the last sentence.

Hateful message: \texttt{{hateful\_message}}

Counter-speech:
\end{tcolorbox}

\noindent In order to use JudgeLM as a metric, we used the following prompts. The first one is used to compare RAG methods against No RAG methods, and against the MT-Co baseline. The second prompt is used to compare our RAG-based CS against those generated with competitive approaches~\cite{wilk2025fact,russo2025trenteam,bonaldi2025first}.

\begin{tcolorbox}[promptbox, title={1. JudgeLM — RAG vs No RAG}]
\itshape
You are an impartial evaluator. You will be given a piece of hate speech \texttt{\{hs\}} and two counter-speech responses \texttt{\{cs\}}. Your task is to judge which counter-speech is better according to the following criteria: 

1) \textbf{Factuality} – Does the response contain accurate and verifiable information? A factually correct response is always preferred.  
2) \textbf{Number of facts} – Does the response provide multiple relevant facts, not just one? More well-chosen facts are better.  
3) \textbf{Relevance} – Does the response directly address the hate speech?  
4) \textbf{Specificity} – Is the response specific and non-generic?
\end{tcolorbox}

\begin{tcolorbox}[promptbox, title={2. JudgeLM — Method Comparison}]
\itshape
You are an impartial evaluator. You will be given a piece of hate speech \texttt{\{hs\}} and two counter-speech responses \texttt{\{cs\}}. Your task is to judge which response is better according to the following criteria:

- The response must be factual, easy to read, empathetic, and suitable as a short social media comment.  
- Friendly and colloquial language is preferred. Responses that sound like essays, lectures, or academic articles should be penalized.  
- Conciseness: The response should be around 2 sentences long. 
\end{tcolorbox}

\paragraph{Details on the paragraph splitting.} The retrieval step is done on the documents of the knowledge base, splitted in paragraphs. To obtain them, each document’s text is split by double newlines \texttt{\textbackslash n\textbackslash n}
to obtain initial paragraphs, which are cleaned by removing extra spaces and replacing newlines with spaces. Paragraphs exceeding a specified maximum token length (150) are further split by sentences using \texttt{NLTK}’s sent\_tokenize, forming smaller chunks that stay under the token limit, considering average paragraph length. The \texttt{GPT-2} tokenizer is used to count tokens in each paragraph, ensuring compliance with the length restriction. Each resulting paragraph is stored as a dictionary containing its text and document ID, and all units are saved to a cache file for fast reuse without repeating the splitting process.

\begin{table*}[h!]
\centering
\footnotesize

\begin{tabular}{l l c c c c}
\toprule
\multicolumn{6}{c}{\textbf{No-RAG vs Retrieval: Per-sample Automatic Metrics}} \\
\midrule
\textbf{Model} & \textbf{Metric} & \textbf{Friedman} & \textbf{BM25} & \textbf{SBERT} & \textbf{BGE} \\
\midrule

\multirow{5}{*}{Llama}
 & BLEU     & $2.0\times10^{-307}$ & $1.1\times10^{-182}$ & $2.3\times10^{-144}$ & $1.4\times10^{-132}$ \\
 & METEOR   & $4.4\times10^{-39}$  & $7.2\times10^{-3}$   & $5.5\times10^{-11}$  & $9.0\times10^{-34}$ \\
 & ROUGE-L  & $1.5\times10^{-33}$  & $9.3\times10^{-37}$  & $7.3\times10^{-18}$  & $8.4\times10^{-9}$ \\
 & BERTScore$_{F1}$ & $<10^{-300}$ & $<10^{-300}$ & $6.8\times10^{-276}$ & $1.8\times10^{-257}$ \\
 & Safety   & $3.6\times10^{-175}$ & $<10^{-79}$ & $<10^{-134}$ & $<10^{-173}$ \\

\midrule
\multirow{5}{*}{CommandR}
 & BLEU     & $<10^{-300}$ & $2.5\times10^{-307}$ & $9.4\times10^{-220}$ & $2.5\times10^{-177}$ \\
 & METEOR   & $2.2\times10^{-102}$ & $4.5\times10^{-37}$ & $1.2\times10^{-90}$ & $8.8\times10^{-79}$ \\
 & ROUGE-L  & $3.5\times10^{-31}$ & $4.9\times10^{-30}$ & $3.2\times10^{-7}$ & $1.5\times10^{-1}$ \\
 & BERTScore$_{F1}$ & $<10^{-300}$ & $<10^{-300}$ & $<10^{-300}$ & $<10^{-300}$ \\
 & Safety   & $<10^{-300}$ & $<10^{-185}$ & $<10^{-300}$ & $<10^{-300}$ \\

\midrule
\multirow{5}{*}{Mistral}
 & BLEU     & $<10^{-300}$ & $4.9\times10^{-243}$ & $2.1\times10^{-194}$ & $6.6\times10^{-124}$ \\
 & METEOR   & $7.1\times10^{-18}$ & $1.8\times10^{-4}$ & $5.4\times10^{-10}$ & $2.3\times10^{-21}$ \\
 & ROUGE-L  & $2.9\times10^{-58}$ & $6.5\times10^{-58}$ & $2.9\times10^{-37}$ & $5.4\times10^{-11}$ \\
 & BERTScore$_{F1}$ & $<10^{-300}$ & $<10^{-300}$ & $<10^{-300}$ & $4.4\times10^{-127}$ \\
 & Safety   & $<10^{-300}$ & $<10^{-217}$ & $<10^{-251}$ & $<10^{-281}$ \\

\midrule
\multirow{5}{*}{GPT}
 & BLEU     & $<10^{-300}$ & $<10^{-300}$ & $5.5\times10^{-299}$ & $3.6\times10^{-226}$ \\
 & METEOR   & $1.9\times10^{-97}$ & $1.6\times10^{-39}$ & $1.5\times10^{-63}$ & $3.4\times10^{-92}$ \\
 & ROUGE-L  & $1.1\times10^{-53}$ & $1.5\times10^{-51}$ & $4.0\times10^{-34}$ & $1.6\times10^{-6}$ \\
 & BERTScore$_{F1}$ & $<10^{-300}$ & $<10^{-300}$ & $<10^{-300}$ & $1.8\times10^{-205}$ \\
 & Safety   & $<10^{-300}$ & $<10^{-207}$ & $<10^{-300}$ & $<10^{-300}$ \\

\midrule
\multicolumn{6}{c}{\textbf{NLI \& Grounding Metrics (Evidence vs Counter-speech)}} \\
\midrule
\textbf{Model} & \textbf{Metric} & \textbf{Friedman} & \textbf{BM25 vs SBERT} & \textbf{BM25 vs BGE} & \textbf{SBERT vs BGE} \\
\midrule

\multirow{4}{*}{Llama}
 & Contradiction & $2.14\times10^{-10}$ & $5.87\times10^{-07}$ & $3.66\times10^{-04}$ & $1.00$ \\
 & Entailment    & $5.93\times10^{-01}$ & $1.00$ & $1.00$ & $1.00$ \\
 & Neutral       & $6.67\times10^{-02}$ & $3.27\times10^{-01}$ & $1.00$ & $3.00\times10^{-02}$ \\
 & Grounding     & $2.41\times10^{-178}$ & $2.66\times10^{-92}$ & $2.04\times10^{-28}$ & $6.09\times10^{-217}$ \\

\midrule
\multirow{4}{*}{CommandR}
 & Contradiction & $1.10\times10^{-07}$ & $1.17\times10^{-08}$ & $8.32\times10^{-02}$ & $2.84\times10^{-03}$ \\
 & Entailment    & $4.12\times10^{-50}$ & $8.66\times10^{-49}$ & $3.88\times10^{-01}$ & $1.00\times10^{-66}$ \\
 & Neutral       & $5.20\times10^{-38}$ & $3.43\times10^{-32}$ & $5.30\times10^{-03}$ & $1.21\times10^{-50}$ \\
 & Grounding     & $2.59\times10^{-241}$ & $2.46\times10^{-142}$ & $4.18\times10^{-34}$ & $2.55\times10^{-302}$ \\

\midrule
\multirow{4}{*}{Mistral}
 & Contradiction & $2.61\times10^{-20}$ & $7.42\times10^{-03}$ & $2.80\times10^{-02}$ & $3.28\times10^{-09}$ \\
 & Entailment    & $3.22\times10^{-20}$ & $1.13\times10^{-28}$ & $4.59\times10^{-02}$ & $2.12\times10^{-43}$ \\
 & Neutral       & $2.91\times10^{-21}$ & $6.04\times10^{-24}$ & $3.14\times10^{-01}$ & $1.54\times10^{-31}$ \\
 & Grounding     & $2.54\times10^{-309}$ & $1.06\times10^{-122}$ & $3.33\times10^{-78}$ & $0.00$ \\

\midrule
\multirow{4}{*}{GPT}
 & Contradiction & $3.53\times10^{-68}$ & $8.20\times10^{-02}$ & $9.61\times10^{-28}$ & $2.36\times10^{-41}$ \\
 & Entailment    & $1.32\times10^{-111}$ & $1.19\times10^{-30}$ & $2.33\times10^{-56}$ & $4.86\times10^{-164}$ \\
 & Neutral       & $2.33\times10^{-94}$ & $1.02\times10^{-38}$ & $4.41\times10^{-39}$ & $4.47\times10^{-147}$ \\
 & Grounding     & $0.00$ & $3.67\times10^{-126}$ & $0.00$ & $0.00$ \\

\bottomrule
\end{tabular}

\caption{Statistical significance results for No-RAG vs retrieval methods for per-sample automatic metrics (top) and 
statistical significance of NLI and grounding metrics between retrieval strategies only, since No-RAG has no retrieved evidence. All values are Bonferroni-corrected Wilcoxon p-values following Friedman tests.}
\label{tab:stats_combined_all_metrics}
\end{table*}

\begin{table*}[t]
\centering
\small
\begin{tabular}{llrrrrrr}
\toprule
\textbf{Model} & \textbf{Setting} & \textbf{Tokens} & \textbf{Est. FLOPs} & \textbf{GPU-time} & \textbf{Energy} & \textbf{CO\textsubscript{2}e} & \textbf{Cost} \\
\midrule
Llama-3.1-8B  & No-RAG            & 585,407    & $9.40\times10^{15}$ & 1.6\,h  & 0.63\,kWh & 0.25\,kg & --      \\
Llama-3.1-8B  & RAG (avg/retriever) & 10,960,856 & $1.76\times10^{17}$ & 42.2\,h & 16.9\,kWh & 6.76\,kg & --      \\
Mistral-7B    & No-RAG            & 533,977    & $7.74\times10^{15}$ & 1.1\,h  & 0.45\,kWh & 0.18\,kg & --      \\
Mistral-7B    & RAG (avg/retriever) & 5,007,558  & $7.26\times10^{16}$ & 6.4\,h  & 2.56\,kWh & 1.02\,kg & --      \\
CommandR-7B   & No-RAG            & 514,684    & $7.21\times10^{15}$ & 1.3\,h  & 0.51\,kWh & 0.20\,kg & --      \\
CommandR-7B   & RAG (avg/retriever) & 5,021,287  & $7.03\times10^{16}$ & 8.3\,h  & 3.19\,kWh & 1.28\,kg & --      \\
GPT-4o-mini   & No-RAG            & 450,233    & --                   & --      & --        & --       & \$0.164 \\
GPT-4o-mini   & RAG (avg/retriever) & 4,818,657  & --                   & --      & --        & --       & \$1.224 \\
\midrule
\textbf{Open-weight total} & \textbf{No-RAG} & \textbf{2,084,301}  & $\mathbf{2.44\times10^{16}}$ & \textbf{4.0\,h}   & \textbf{1.59\,kWh} & \textbf{0.64\,kg} & --      \\
\textbf{Open-weight total} & \textbf{RAG (3 retrievers)} & \textbf{77,425,075} & $\mathbf{9.57\times10^{17}}$ & \textbf{133.3\,h} & \textbf{53.3\,kWh} & \textbf{21.3\,kg} & --      \\
\bottomrule
\end{tabular}
\caption{Estimated computational cost of RAG vs.\ No-RAG CS generation. RAG results are averaged per retriever (BM25, SBERT, BGE-M3) unless marked as totals across all three.}
\label{tab:compute_cost}
\end{table*}

\paragraph{Deployment feasibility and computational cost.} All summarization and generation experiments were conducted on an A100 GPU, with the following parameters: \texttt{max\_new\_tokens}=150, and \texttt{temperature}=0.5. Because wall-clock timing was 
not logged during the original experiments, we estimate computational cost post-hoc by reconstructing the exact prompts sent to each model from our saved generation logs and tokenizing them with \texttt{tiktoken} (GPT-4o-mini's \texttt{o200k\_base} encoder). For GPT-4o-mini, token counts are converted to API costs using its published per-token pricing (\$0.15 / \$0.60 per 1M input/output tokens). Across all summarization and generation tasks, for RAG with the three retrievers and No-RAG, over 5,003 HS messages, this reconstruction yields 53.8M input tokens and 25.7M output tokens in total.
RAG's quality gains come with extra computational and deployment costs. We estimate the number of Floating-Point Operations (FLOPs) using the standard formula $2 \times N_{\text{params}} \times \text{tokens}$. GPU time is estimated by dividing the number of output tokens by each model's published decoding speed on a single A100 GPU. From this time, we estimate energy use (GPU time $\times$ the A100's rated power draw) and CO\textsubscript{2}e emissions (energy $\times$ an assumed grid carbon-intensity factor).\footnote{We assume the A100's rated 400W thermal design power (NVIDIA A100 datasheet, \url{https://www.nvidia.com/en-us/data-center/a100/}) and a grid carbon-intensity factor of 0.4\,kg CO\textsubscript{2}/kWh, a common placeholder in ML carbon-accounting work, given that measured intensities vary widely by region, from roughly 20 to over 700\,g CO\textsubscript{2}/kWh (Lacoste et al., 2019, arXiv:1910.09700).} We note that GPU time, energy, and CO\textsubscript{2}e are rougher estimates than FLOPs or GPT-4o-mini's cost: FLOPs follow directly from token and parameter counts, and GPT-4o-mini's cost uses real, published pricing, but GPU time, energy, and CO\textsubscript{2}e also depend on assumptions about decoding speed, hardware power draw, and grid carbon intensity, all of which vary in practice. Retrieval cost is not included, as it is small compared to LLM inference. Across the three open-weight models, RAG needs 33-39$\times$ more compute than No-RAG (Table~\ref{tab:compute_cost}), mainly because each RAG generation also requires three summarization calls ($k=3$). For GPT-4o-mini, generating CS for all 5,003 HS messages with all three retrievers costs an estimated \$3.67 in API fees, compared to \$0.16 without RAG. Self-hosting an open-weight model avoids relying on a third-party API, but costs more in infrastructure, in fact running the full open-weight RAG experiments took an estimated 133 GPU-hours on a single A100. This points to a trade-off for real deployment: using an API keeps costs low but raises questions about data governance and dependence on an external provider, while self-hosting gives more control but is currently too slow for real-time, large scale use without further optimization.

\section{Statistical Significance of Retrieval Effects}
\label{app:stat_sig}

We conduct non-parametric Friedman tests followed by Bonferroni-corrected Wilcoxon signed-rank tests to evaluate whether RAG retrieval strategies (BM25, Sentence-BERT, BGE) significantly affect model outputs compared to the No-RAG baseline for the per-sample automatic metrics (BLEU, METEOR, ROUGE-L, BERTScore, Safety), which allow direct pairwise comparison across systems. In contrast, diversity metrics such as Distinct-1, Distinct-2, and Repetition Rate are computed at the corpus level, producing a single value per model. Because these measures do not yield per-sample scores and thus lack within-system variance, statistical testing is not applicable. We also perform the same tests for the per-sample NLI metrics between retrieval strategies only, since No-RAG models have no retrieved evidence. Table~\ref{tab:stats_combined_all_metrics} shows that, for all models, Friedman tests reveal significant overall effects ($p < 0.001$) for the standard quality and safety metrics, and pairwise comparisons confirm that retrieval methods consistently produce statistically significant differences. For the NLI metrics, extremely small $p$-values for most scores indicate that the choice of retriever can significantly impact CS grounding and hallucination mitigation, although some metrics (e.g., evidence entailment for certain models) show weaker or non-significant effects.


\section{Participants' Demographic Characteristics}
\label{app:demo}

\begin{figure}[h]
\centering
\includegraphics[width=0.4\textwidth]{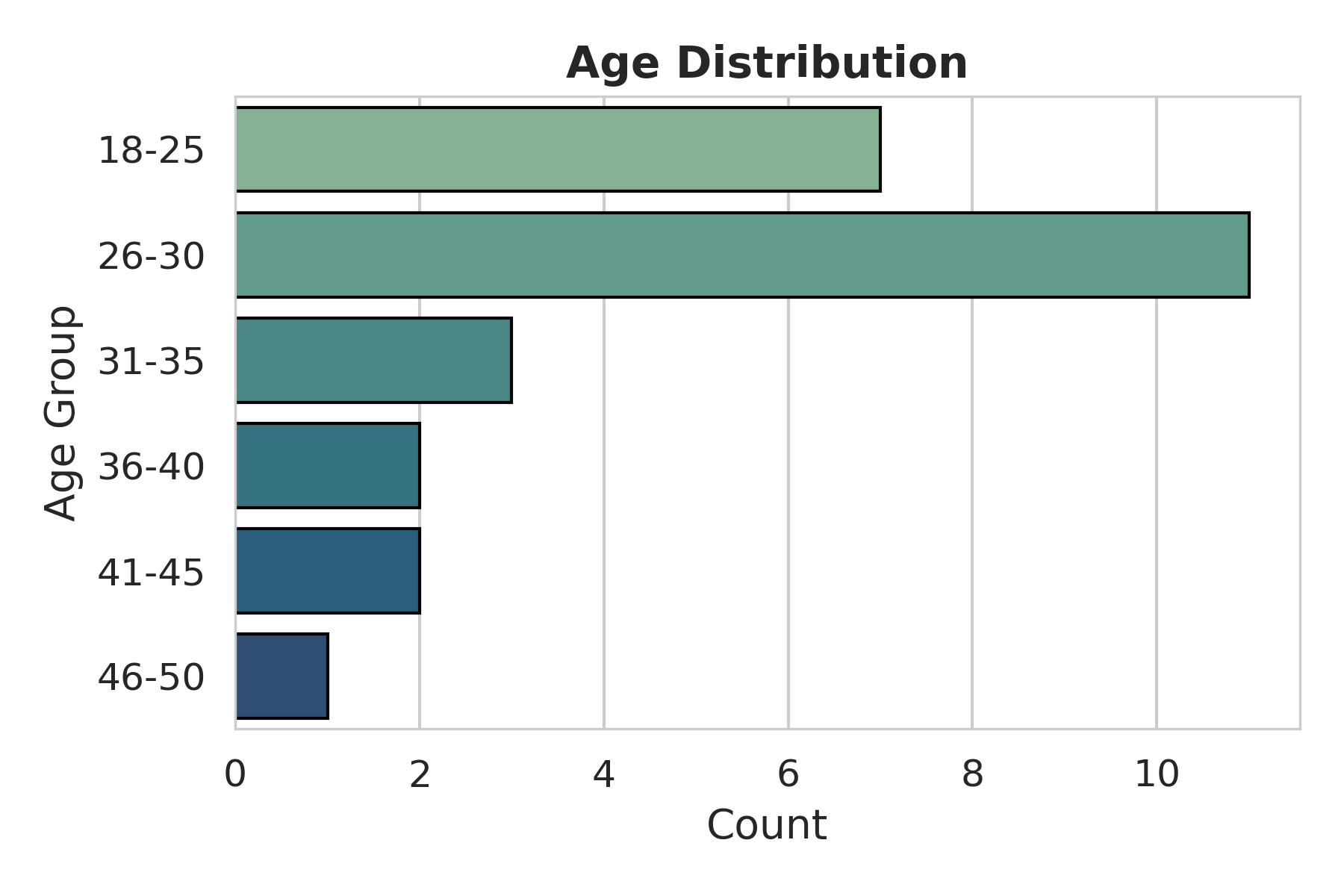}
\caption{Distribution of age of the 26 participants.}
\label{fig:age}
\end{figure}

\begin{figure}[h]g
\centering
\includegraphics[width=0.4\textwidth]{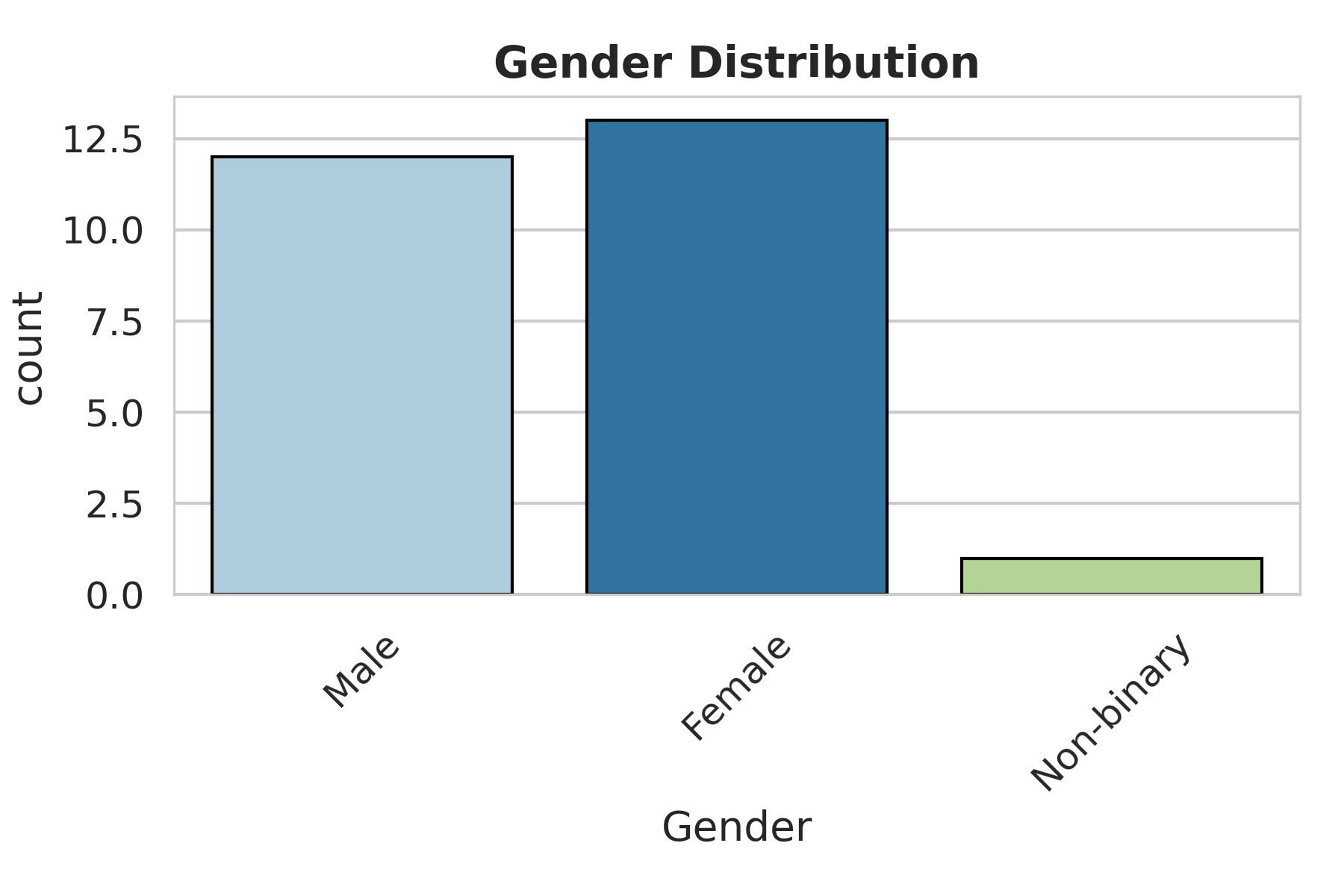}
\caption{Distribution of gender of the 26 participants.}
\label{fig:gender}
\end{figure}

\begin{figure}[h]
\centering
\includegraphics[width=0.4\textwidth]{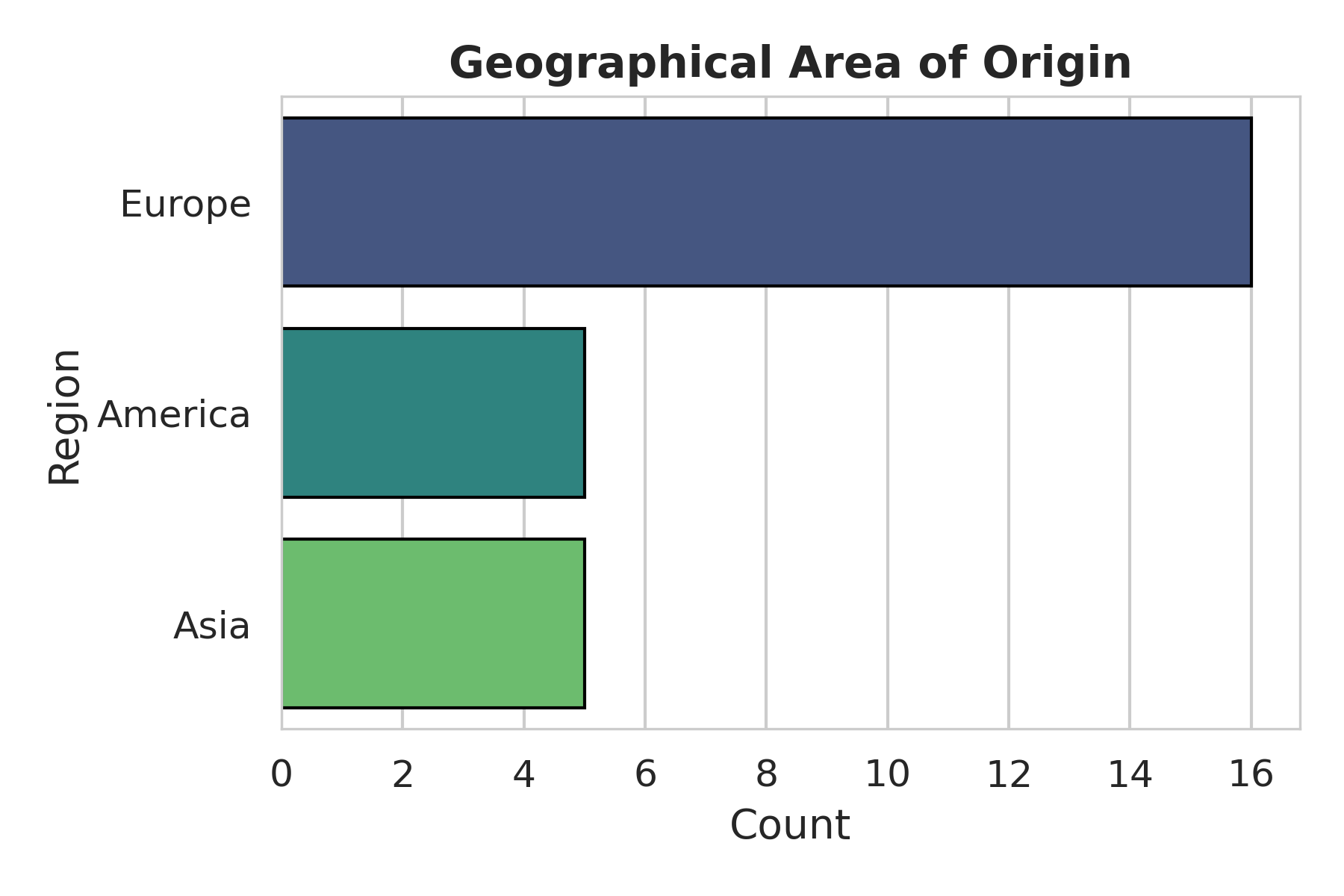}
\caption{Distribution of the geographical area of origin of the 26 participants.}
\label{fig:geo}
\end{figure}

\begin{figure}[h]
\centering
\includegraphics[width=0.4\textwidth]{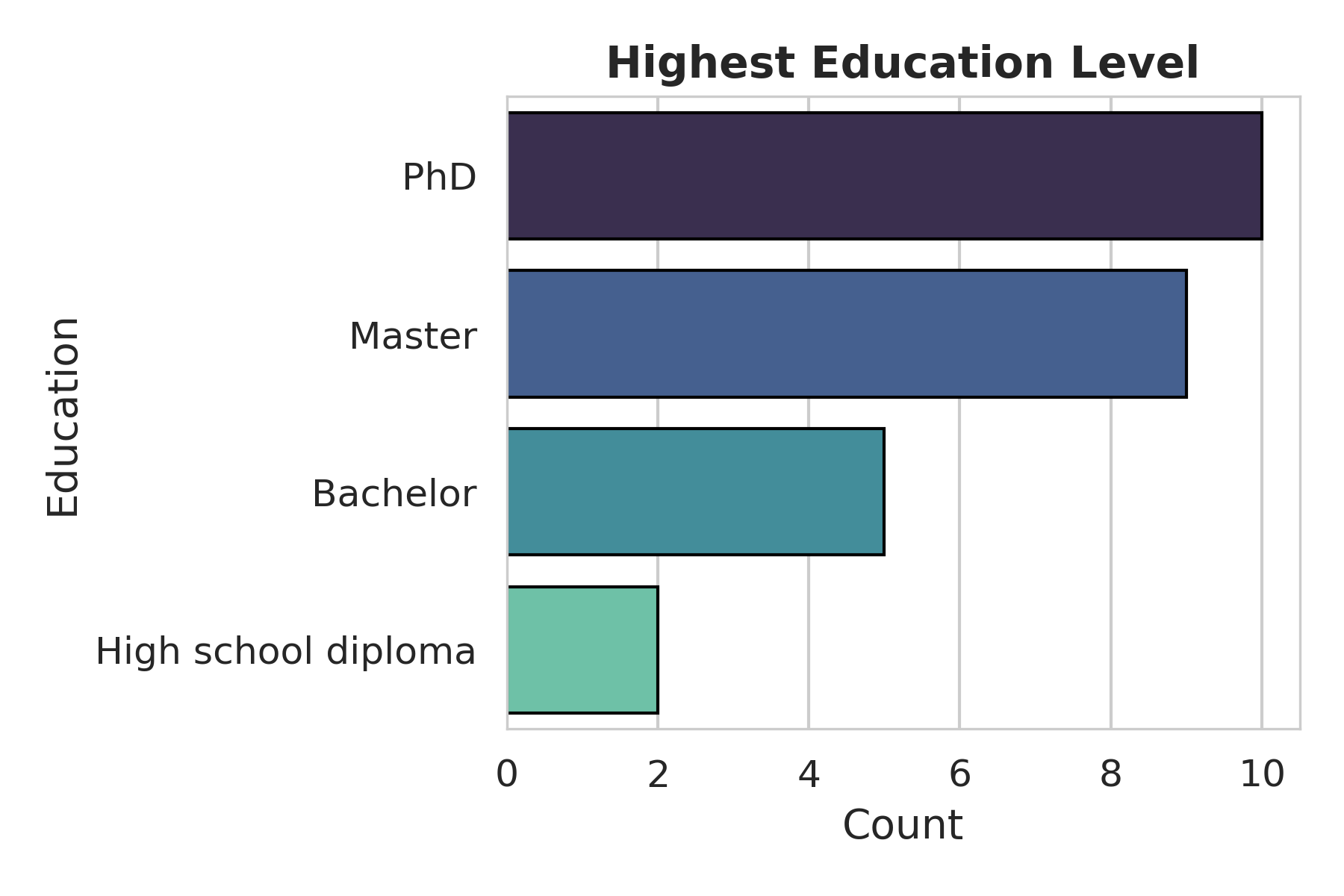}
\caption{Distribution of level of education of the 26 participants.}
\label{fig:edu}
\end{figure}

In this section, we report the demographic characteristics of the 26 participants of the human evaluation, which were recruited on a voluntary basis.
Figures \ref{fig:age}, \ref{fig:gender}, \ref{fig:geo}, \ref{fig:edu} show the distribution of age groups, gender, geographical area of origin, and the highest obtained education level. Age varies between 18 and 50 years, with the majority group being between 18 and 35. Gender is evenly distributed among females and males with one person identifying as non-binary. The geographical area of origin covers all major areas, including Europe, North/South America, and Asia. Education level spans from high school diploma to PhD, with the majority of respondents having pursued a PhD. We also asked the respondents their area of expertise, which covers the following fields: Computer Science, Computer Engineering, Data Science, Psychology, Natural Language Processing, Linguistics, and Management.

\begin{figure}[h]
\centering
\includegraphics[width=0.5\textwidth]{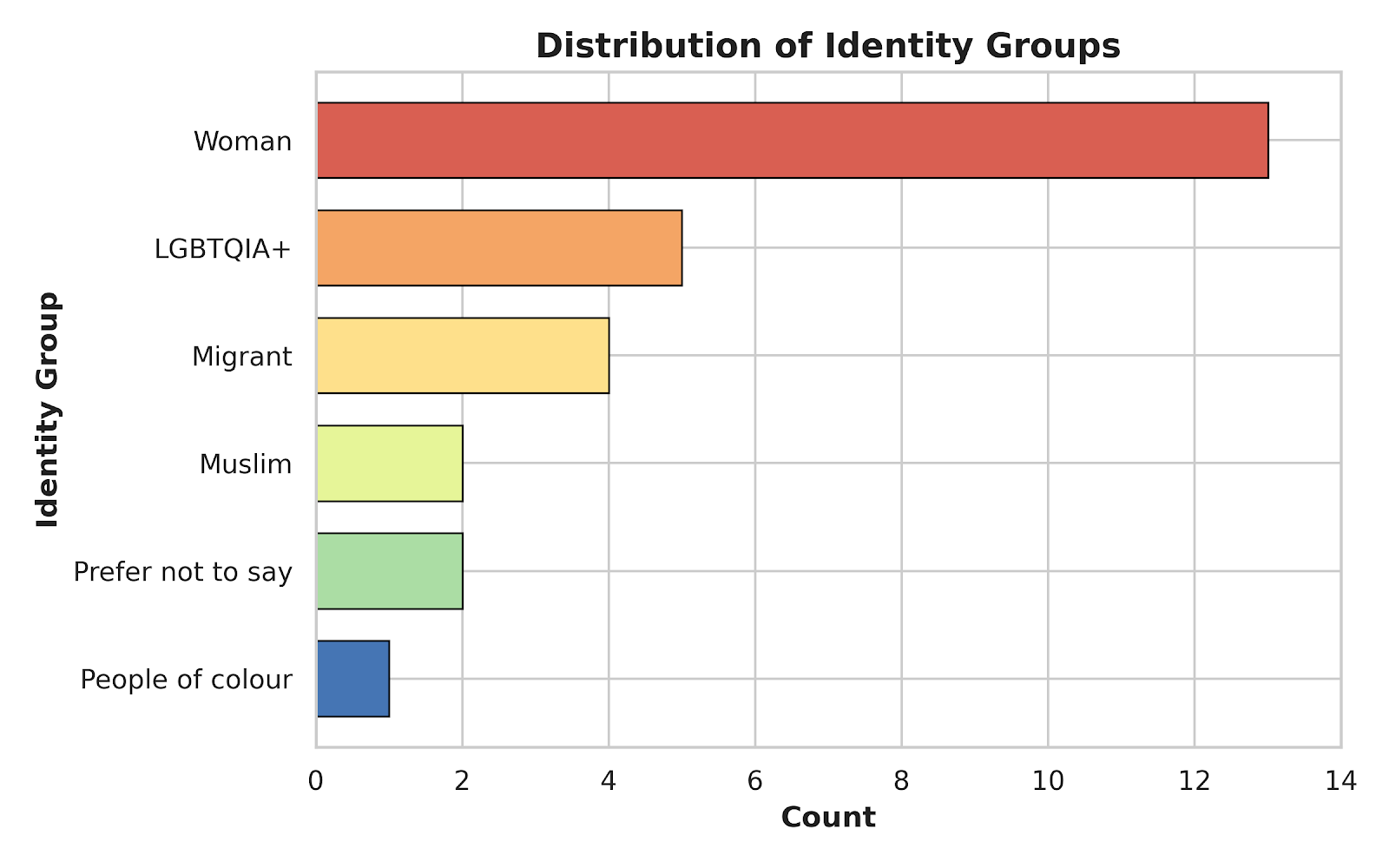}
\caption{Distribution of the target group identification of the 26 participants.}
\label{fig:target}
\end{figure}

Figure \ref{fig:target} shows the target groups in which the respondents identify. They belong to almost all the HS targets we consider in our analysis, with the majority of them identifying as women.

\section{Guidelines for Human Evaluation} \label{app:guidelines}

As described in Section 
8 of the main paper, we carried out a human evaluation. Participants were voluntarily recruited and could opt out from the study at any time. Precise instructions were given, highlighting the potential risks and distress of the study. We define the following metrics used for the human evaluation: Relevance, Factuality, Cogency, and Correctness use a Likert scale from 1 to 3 (with 3 being the best possible score), while Effectiveness, and Is the Best are binary dimensions to which the participants could reply "Yes" or "No". They are defined as follows:

\paragraph{Relevance.} How relevant is the counter-speech to the hate speech, in terms of topic and target of hate (i.e. the offended minority)?
\begin{etaremune}
\item The CS addresses both the correct HS target and the topic.
\item The CS addresses correctly only the topic or the target of hate.
\item The CS is very general: the same message could reply to whatever HS.
\end{etaremune} 

\paragraph{Factuality.} How informative is the counter-speech as a response to its hate speech, in terms of quantity and factual correctness of included facts?
\begin{etaremune}
\item The CS provides multiple logically correct arguments and they are all sound and relevant.
\item The CS provides only one logically correct argument which is sound and relevant.
\item No reasons are provided for the CS claim, or none of the reasons are relevant to or support the CS claim.
\end{etaremune} 

\paragraph{Cogency.} This dimension measures the quantity of the supporting logically correct arguments provided by the counter-speech.
\begin{etaremune}
\item The CS provides multiple information not present in the HS, and they are all factually correct.
\item The CS provides only one information not present in the HS, and it is correct.
\item The CS provides no additional information with respect to the HS.
\end{etaremune} 

\paragraph{Correctness.} How much is the style of the CS correct? I.e., the CS is free of grammatical and syntactical errors and it is not hateful (does not contain toxic language).
\begin{etaremune}
\item The CS has not grammatical or syntactical errors AND does not have toxic language.
\item The CS has grammatical and syntactical errors OR it has a toxic language.
\item The CS has grammatical or syntactical errors AND has a toxic language.
\end{etaremune} 

\paragraph{Effectiveness.} Is the counter-speech likely to be persuasive and change someone’s perspective on the issue? Is the CS able to change the opinion of the author of the HS?

\paragraph{Is the Best?} Refers to whether the counter-speech is the best one among the ones generated for the same HS. Select the CS that you prefer based on the scores you provided.

\noindent We also provided the following examples to better understand the metrics' scores.

\mdfdefinestyle{hsframe}{
  backgroundcolor=white, 
  linecolor=gray!60,
  roundcorner=5pt,
  innertopmargin=5pt,
  innerbottommargin=5pt,
  leftmargin=0pt,
  rightmargin=0pt
}

\mdfdefinestyle{csframe}{
  backgroundcolor=white,
  linecolor=gray!40,
  roundcorner=3pt,
  innertopmargin=4pt,
  innerbottommargin=4pt,
  leftmargin=5pt,
  rightmargin=5pt
}

\begin{mdframed}[style=hsframe]
\textbf{HS example:} “Immigrants are lazy and just come here to steal jobs.”
\end{mdframed}

\begin{mdframed}[style=csframe]
\textbf{1. Relevance}\\[2pt]
\textit{Score 3 (Topic + Target correct):}\\
“That’s not true—immigrants contribute greatly to the economy by working hard in sectors like healthcare and construction.”\\[4pt]

\textit{Score 2 (Only topic or target correct):}\\
“Jobs are important for everyone, and we should all value hard work.” \textit{(Topic correct, but not addressing immigrants directly.)}\\
“Immigrants deserve respect.” \textit{(Target correct, but no engagement with the specific topic of jobs/laziness.)}\\[4pt]

\textit{Score 1 (Very general):}\\
“We should all just be kind to each other.”
\end{mdframed}

\begin{mdframed}[style=csframe]
\textbf{2. Correctness}\\[2pt]
\textit{Score 3 (No errors, no toxicity):}\\
“Immigrants are not lazy—many work multiple jobs to support their families.”\\[4pt]

\textit{Score 2 (Some errors OR mild toxicity, but not both):}\\
“Immigrants isn’t lazy, they works a lot.” \textit{(Grammatical errors but not offensive.)}\\
“That’s a dumb thing to say, immigrants work hard.” \textit{(Toxic tone but grammatically correct.)}\\[4pt]

\textit{Score 1 (Errors AND toxic):}\\
“U r idiot, immigrants aren’t ur slaves.”
\end{mdframed}

\begin{mdframed}[style=csframe]
\textbf{3. Factuality}\\[2pt]
\textit{Score 3 (Multiple correct facts):}\\
“Immigrants make up 17\% of the healthcare workforce in the US and many work in agriculture, which keeps food affordable.”\\[4pt]

\textit{Score 2 (One correct fact):}\\
“Studies show that immigrants often take jobs that citizens don’t want, like farm labor.”\\[4pt]

\textit{Score 1 (No new info):}\\
“Immigrants work hard.” \textit{(Assertion without factual support.)}
\end{mdframed}

\begin{mdframed}[style=csframe]
\textbf{4. Cogency}\\[2pt]
\textit{Score 3 (Multiple sound reasons):}\\
“It’s unfair to call immigrants lazy—many work multiple jobs, they often contribute more in taxes than they receive in benefits, and they fill critical roles in the economy.”\\[4pt]

\textit{Score 2 (One sound reason):}\\
“It’s unfair to call immigrants lazy because many of them work long hours.”\\[4pt]

\textit{Score 1 (No or irrelevant reasons):}\\
“Immigrants are great, stop hating.” \textit{(Claim, but no logical support.)}
\end{mdframed}


\section{Full Results of the Per-Target Human Evaluation}
\label{appendix:full_stats}

In this section, we report the full mixed-effects model results for the four quality metrics, along with disaggregated per-target results on the same metrics and effectiveness rates for the human evaluation analyses described in Section 8 of the main paper.

\subsection{Mixed-Effects Model Results}
 
To analyze how CS quality ratings vary by target group and CS strategy while accounting for message-level and annotator-level variability, we estimate the following mixed-effects model, fitted separately for each quality dimension (Relevance, Factuality, Cogency, Correctness):

\begin{align}
\text{Rating}_{ijkl} &= \beta_0 
+ \beta_1 \text{TargetGroup}_i 
+ \beta_2 \text{CSStrategy}_j \nonumber \\
&\quad + \beta_3 (\text{TargetGroup} \times \text{CSStrategy})_{ij} \nonumber \\
&\quad + u_k^{\text{HS}} 
+ u_l^{\text{Annotator}}
+ \epsilon_{ijkl}
\end{align}

\noindent where:

\begin{itemize}
    \item $\text{Rating}_{ijkl}$ is the score given by annotator $l$ for CS strategy $j$ targeting HS instance $k$ for target group $i$.
    \item $\text{TargetGroup}_i$ is a binary indicator for whether the annotator identifies with the target group.
    \item $\text{CSStrategy}_j$ indexes the CS strategy used.
    \item $(\text{TargetGroup} \times \text{CSStrategy})_{ij}$ is the interaction term capturing whether in-group vs out-group raters evaluate specific strategies differently.
    \item $u_k^{\text{HS}}$ and $u_l^{\text{Annotator}}$ are random intercepts for each HS and annotator.
    \item $\epsilon_{ijkl}$ is the residual error term.
\end{itemize}

\paragraph{Identity main effects (Model 1).} 
Across all dimensions, in-group participants assign significantly lower ratings than out-group participants when considered in isolation:
\begin{itemize}
    \item Relevance: $\beta=-0.212$, $p<0.001$  
    \item Factuality: $\beta=-0.292$, $p<0.001$  
    \item Cogency: $\beta=-0.128$, $p<0.001$  
    \item Correctness: $\beta=-0.339$, $p<0.001$
\end{itemize}

\paragraph{Controlling for CS and message variability (Model 2).}  
When including CS and message-level random effects, the main effect of identity becomes non-significant across all dimensions, indicating that differences in perceived quality are largely driven by CS rather than identity alone. CS strategies consistently predict higher ratings, with all coefficients significant ($p<0.001$). Message-level variance is non-trivial (3.3–5.8), reflecting heterogeneity across HS messages.

\paragraph{Identity × CS interactions (Model 3).}  
Interactions reveal that in-group participants are more selective, assigning lower ratings for specific strategies (i.e., majority of No-RAG strategies):  

\begin{itemize}
    \item \textbf{Relevance}: Llama No RAG, CommandR No RAG, Mistral No RAG, GPT No RAG
    \item \textbf{Factuality}: Llama No RAG, Llama RAG, CommandR No RAG, CommandR RAG, Mistral No RAG, Mistral RAG, GPT No RAG, GPT RAG
    \item \textbf{Cogency}: Llama No RAG, CommandR No RAG, Mistral RAG, GPT No RAG
    \item \textbf{Correctness}: CommandR No RAG, Mistral No RAG, GPT No RAG
\end{itemize}

Main effects of identity remain non-significant, confirming that selectivity is strategy-specific rather than uniform.

\subsection{Disaggregated Results}

Table~\ref{tab:cs_combined_full} reports CS evaluation results by LLM, RAG status, and metric for in-group and out-group raters. Panel A reports mean human ratings (1-3) for Relevance, Factuality, Cogency, and Correctness, disaggregated by LLM and RAG status, with standard deviations. Panel B reports mean effectiveness rates (0-1) of counter-speech strategies, with standard deviations. Panel C shows aggregate strategy diversity metrics. There is a clear trend where RAG-generated CS obtains better scores than its No-RAG counterparts, across all quality metrics and effectiveness rates. Finally, Panel~C shows that in-group participants exhibit lower entropy in strategy use, and higher concentration on a small subset of preferred strategies.
\begin{table}[!t]
\centering
\footnotesize
\begin{tabular}{llcc}
\toprule
\textbf{LLM / RAG} & \textbf{Metric} & \textbf{In-group} & \textbf{Out-group} \\
\midrule
CommandR No RAG & Cogency      & 2.44 $\pm$ 0.59 & 2.17 $\pm$ 0.74 \\
                & Correctness  & 2.91 $\pm$ 0.28 & 2.67 $\pm$ 0.66 \\
                & Factuality   & 2.05 $\pm$ 0.65 & 1.76 $\pm$ 0.77 \\
                & Relevance    & 2.83 $\pm$ 0.41 & 2.55 $\pm$ 0.68 \\
CommandR RAG    & Cogency      & 2.51 $\pm$ 0.55 & 2.46 $\pm$ 0.66 \\
                & Correctness  & 2.81 $\pm$ 0.42 & 2.71 $\pm$ 0.62 \\
                & Factuality   & 2.44 $\pm$ 0.65 & 2.17 $\pm$ 0.76 \\
                & Relevance    & 2.90 $\pm$ 0.30 & 2.68 $\pm$ 0.62 \\
GPT No RAG      & Cogency      & 2.50 $\pm$ 0.61 & 2.27 $\pm$ 0.69 \\
                & Correctness  & 2.90 $\pm$ 0.38 & 2.65 $\pm$ 0.67 \\
                & Factuality   & 2.22 $\pm$ 0.63 & 1.77 $\pm$ 0.76 \\
                & Relevance    & 2.83 $\pm$ 0.41 & 2.53 $\pm$ 0.65 \\
GPT RAG         & Cogency      & 2.66 $\pm$ 0.55 & 2.49 $\pm$ 0.71 \\
                & Correctness  & 2.88 $\pm$ 0.37 & 2.70 $\pm$ 0.62 \\
                & Factuality   & 2.68 $\pm$ 0.55 & 2.29 $\pm$ 0.77 \\
                & Relevance    & 2.83 $\pm$ 0.44 & 2.61 $\pm$ 0.67 \\
Llama No RAG    & Cogency      & 2.73 $\pm$ 0.45 & 2.39 $\pm$ 0.70 \\
                & Correctness  & 2.89 $\pm$ 0.39 & 2.72 $\pm$ 0.62 \\
                & Factuality   & 2.43 $\pm$ 0.63 & 2.01 $\pm$ 0.81 \\
                & Relevance    & 2.91 $\pm$ 0.33 & 2.63 $\pm$ 0.65 \\
Llama RAG       & Cogency      & 2.64 $\pm$ 0.60 & 2.48 $\pm$ 0.72 \\
                & Correctness  & 2.84 $\pm$ 0.46 & 2.71 $\pm$ 0.61 \\
                & Factuality   & 2.63 $\pm$ 0.58 & 2.45 $\pm$ 0.69 \\
                & Relevance    & 2.75 $\pm$ 0.54 & 2.62 $\pm$ 0.68 \\
MT-Co No RAG    & Cogency      & 1.75 $\pm$ 0.70 & 1.63 $\pm$ 0.75 \\
                & Correctness  & 2.45 $\pm$ 0.81 & 2.20 $\pm$ 0.82 \\
                & Factuality   & 1.44 $\pm$ 0.67 & 1.36 $\pm$ 0.67 \\
                & Relevance    & 2.18 $\pm$ 0.73 & 2.02 $\pm$ 0.74 \\
Mistral No RAG  & Cogency      & 2.20 $\pm$ 0.66 & 2.07 $\pm$ 0.78 \\
                & Correctness  & 2.75 $\pm$ 0.52 & 2.67 $\pm$ 0.52 \\
                & Factuality   & 1.94 $\pm$ 0.70 & 1.69 $\pm$ 0.79 \\
                & Relevance    & 2.80 $\pm$ 0.46 & 2.49 $\pm$ 0.72 \\
Mistral RAG     & Cogency      & 2.35 $\pm$ 0.64 & 2.28 $\pm$ 0.75 \\
                & Correctness  & 2.66 $\pm$ 0.66 & 2.71 $\pm$ 0.64 \\
                & Factuality   & 2.25 $\pm$ 0.65 & 2.12 $\pm$ 0.71 \\
                & Relevance    & 2.73 $\pm$ 0.53 & 2.54 $\pm$ 0.71 \\
\midrule
\multicolumn{4}{l}{\textbf{Panel B: Mean Effectiveness Rates by CS strategy}} \\
\midrule
\multicolumn{2}{l}{\textbf{CS Strategy}} & \textbf{In-group} & \textbf{Out-group} \\
\midrule
\multicolumn{2}{l}{CommandR No RAG} & 0.43 $\pm$ 0.10 & 0.53 $\pm$ 0.21 \\
\multicolumn{2}{l}{CommandR RAG}    & 0.64 $\pm$ 0.15 & 0.70 $\pm$ 0.20 \\
\multicolumn{2}{l}{GPT No RAG}      & 0.51 $\pm$ 0.13 & 0.71 $\pm$ 0.23 \\
\multicolumn{2}{l}{GPT RAG}         & 0.71 $\pm$ 0.19 & 0.80 $\pm$ 0.27 \\
\multicolumn{2}{l}{Llama No RAG}    & 0.61 $\pm$ 0.18 & 0.78 $\pm$ 0.20 \\
\multicolumn{2}{l}{Llama RAG}       & 0.75 $\pm$ 0.20 & 0.78 $\pm$ 0.28 \\
\multicolumn{2}{l}{MT-Co}           & 0.18 $\pm$ 0.15 & 0.19 $\pm$ 0.20 \\
\multicolumn{2}{l}{Mistral No RAG}  & 0.40 $\pm$ 0.13 & 0.46 $\pm$ 0.22 \\
\multicolumn{2}{l}{Mistral RAG}     & 0.61 $\pm$ 0.14 & 0.58 $\pm$ 0.36 \\
\midrule
\multicolumn{4}{l}{\textbf{Panel C: Strategy diversity metrics}} \\
\midrule
\multicolumn{2}{l}{\textbf{Metric}} & \textbf{In-group} & \textbf{Out-group} \\
\midrule
\multicolumn{2}{l}{Mean Effectiveness Rate} & 0.53 & 0.69 \\
\multicolumn{2}{l}{Entropy (CS diversity)}  & 0.93 & 1.56 \\
\multicolumn{2}{l}{Top CS proportion}       & 0.62 & 0.40 \\
\bottomrule
\end{tabular}
\caption{CS evaluation results by LLM, RAG status, and metric for in-group and out-group raters.}
\label{tab:cs_combined_full}
\end{table}

\begin{table}[t] \centering \scriptsize \setlength{\tabcolsep}{3pt} \begin{tabular}{llrrrrrr} \toprule \textbf{Setting} & \textbf{Metric} & \textbf{Rel.} & \textbf{Fact.} & \textbf{Cog.} & \textbf{Corr.} & \textbf{max $|\rho|$} & \textbf{Holm $p$} \\ \midrule \multirow{4}{*}{No-RAG} & BLEU & .240 & .056 & .188 & \textbf{.362}$^{*}$ & .362 & .125 \\ & METEOR & .002 & -.014 & .118 & .058 & .118 & 1.000 \\ & ROUGE-L & .076 & .011 & .154 & .216 & .216 & 1.000 \\ & BERTScore & .032 & -.106 & -.002 & .151 & .151 & 1.000 \\ \midrule \multirow{4}{*}{RAG} & BLEU & .255 & -.210 & -.038 & -.131 & .255 & 1.000 \\ & METEOR & .034 & .185 & .132 & -.085 & .185 & 1.000 \\ & ROUGE-L & .197 & .179 & .172 & -.067 & .197 & 1.000 \\ & BERTScore & .019 & .022 & -.071 & -.053 & .071 & 1.000 \\ \midrule \multirow{4}{*}{All systems} & BLEU & .120 & \textbf{-.257}$^{*}$ & -.062 & .113 & .257 & .280 \\ & METEOR & .065 & .085 & .143 & .004 & .143 & 1.000 \\ & ROUGE-L & .150 & .025 & .131 & .090 & .150 & 1.000 \\ & BERTScore & .031 & -.107 & -.068 & .045 & .107 & 1.000 \\ \bottomrule \end{tabular} \caption{Spearman correlations ($\rho$) between automatic metrics and human evaluation dimensions. $^{*}$ indicates $p<.05$ before Holm correction.} \label{tab:correlations}\end{table}

\subsection{Correlation between Automatic and Human Evaluation}

To further investigate the relationship between automatic and human evaluation, we computed Spearman rank correlations between the automatic metrics and human evaluation scores at the HS$\times$CS-instance level. For each instance, human ratings from the 26 annotators were averaged before being matched to the corresponding automatically evaluated CS. We considered four automatic metrics (BLEU, METEOR, ROUGE-L, and BERTScore-F1) and four human evaluation dimensions (Relevance, Factuality, Cogency, and Correctness). Correlations were computed separately for No-RAG, RAG, and all systems combined. Statistical significance was assessed using two-sided tests, with Holm correction applied to account for multiple comparisons.
Table~\ref{tab:correlations} reports the resulting Spearman correlations. Several moderate associations can be observed, although none remains statistically significant after Holm correction. In the No-RAG setting, the strongest association is between BLEU and Correctness ($\rho=.362$, uncorrected $p=.008$), while in the RAG setting the strongest association is between BLEU and Relevance ($\rho=.255$, $p=.123$). When all systems are considered together, BLEU and Factuality show the strongest association ($\rho=-.257$, uncorrected $p=.014$). The two nominally significant correlations do not survive Holm correction ($p_{\mathrm{Holm}}=.125$ and .280, respectively).
Overall, these results provide no statistically significant evidence of a direct correspondence between the automatic metrics and the human evaluation dimensions considered here. At the same time, the presence of moderate associations suggests that the two evaluation approaches may capture partially overlapping but complementary aspects of CS quality. The relatively small number of HS$\times$CS instances, particularly when considering RAG and No-RAG separately, may also limit the statistical power to detect associations.

\begin{table}[ht]
\centering
\scriptsize
\setlength{\tabcolsep}{3pt}
\begin{tabular}{lccccc}
\toprule
\textbf{Target Group} & \textbf{UN-Only (\%)} & \textbf{EU-Only (\%)} & \textbf{Hybrid (\%)} & \textbf{Total EU (\%)} \\
\midrule
\textbf{DISABLED} & 90.00 & 0.00  & 10.00 & 10.00 \\
\textbf{JEWS}     & 70.00 & 10.00 & 20.00 & 30.00 \\
\textbf{LGBT+}    & 50.00 & 10.00 & 40.00 & 50.00 \\
\textbf{MIGRANTS} & 70.00 & 0.00  & 30.00 & 30.00 \\
\textbf{MUSLIMS}  & 0.00  & 40.00 & 60.00 & \textbf{100.00} \\
\textbf{POC}      & 90.00 & 0.00  & 10.00 & 10.00 \\
\textbf{WOMEN}    & 100.00 & 0.00 & 0.00  & 0.00 \\
\textbf{Other}    & 40.00 & 0.00  & 60.00 & 60.00 \\
\midrule
\textbf{Overall / Total} & \textbf{63.75} & \textbf{7.50} & \textbf{28.75} & \textbf{36.25} \\
\bottomrule
\end{tabular}
\caption{Knowledge source retrieval distribution per target identity group in the ablation study subset ($N=80$). ``Total EU'' represents the combined proportion of EU-Only and Hybrid retrievals.}
\label{tab:appendix_source_breakdown}
\end{table}

\section{Additional Ablation Results}
\label{app:ablation}

To evaluate how international and regional policy knowledge bases interact within our framework, we analyzed the top-3 retrieved documents mapped to their originating corpora (UN vs EU) across the ablation study evaluation set (N=80, stratified equally across 8 target groups). As detailed in Table \ref{tab:appendix_source_breakdown}, across the entire subset, 63.75\% of instances retrieve UN-only documents, 7.50\% retrieve EU-only documents, and 28.75\% retrieve a hybrid combination of both, bringing total European legal framework involvement to 36.25\%. Multi-source retrieval exhibits high domain complementarity: queries targeting anti-Muslim HS, for example, rely on EU frameworks in 100\% of cases (40\% EU-Only, 60\% Hybrid) due to specialized European documentation on regional Islamophobia. Conversely, broader identity categories like women (100\% UN) and disabled individuals (90\% UN) draw primarily from global international instruments (e.g., CEDAW, CRPD). Furthermore, we compute context grounding\footnote{Calculated as the token-level precision (unigram overlap) of the generated CS relative to the concatenated top-3 retrieved summaries: $\frac{|\mathcal{T}_{\text{CS}} \cap \mathcal{T}_{\text{Context}}|}{|\mathcal{T}_{\text{CS}}|}$, where higher scores indicate a higher density of directly grounded terminology from the retrieved KB.}. EU-supported generations maintain strong faithfulness, achieving an average context grounding score of 48.51\% (compared to 46.52\% for Hybrid and 43.34\% for UN-Only).

\section{Examples of Generated Counter-Speech} \label{app:exs}

Here, we provide different examples of counter-speech addressing the same hateful message from MT-CONAN. We show CS from GPT No RAG and GPT RAG with BGE-M3.

\begin{mdframed}[linewidth=0.6pt, roundcorner=3pt, innerleftmargin=8pt, innerrightmargin=8pt, innertopmargin=6pt, innerbottommargin=6pt, backgroundcolor=white]
\textbf{HS}: Their power to have movies banned and directors censured is proof that the Jews control Hollywood.
\end{mdframed}

\begin{mdframed}[linewidth=0.6pt, roundcorner=3pt, innerleftmargin=8pt, innerrightmargin=8pt, innertopmargin=6pt, innerbottommargin=6pt, backgroundcolor=white]
\textbf{MT-Co CS}: The ultra rich people in this industry - Jewish or not - have the power and influence. Not sure where you got your facts from regarding Jewish people "controlling" Hollywood.
\end{mdframed}

\begin{mdframed}[linewidth=0.6pt, roundcorner=3pt, innerleftmargin=8pt, innerrightmargin=8pt, innertopmargin=6pt, innerbottommargin=6pt, backgroundcolor=white]
\textbf{GPT No RAG CS}: Attributing the actions of a few individuals to an entire group perpetuates harmful stereotypes and ignores the diverse voices that contribute to the film industry. Let's focus on promoting understanding and celebrating the rich tapestry of talent in Hollywood, rather than spreading division.
\end{mdframed}

\begin{mdframed}[linewidth=0.6pt, roundcorner=3pt, innerleftmargin=8pt, innerrightmargin=8pt, innertopmargin=6pt, innerbottommargin=6pt, backgroundcolor=white]
\textbf{GPT RAG (BGE-M3) CS}: It's important to recognize that attributing the actions of individuals or organizations in Hollywood to an entire group, such as Jews, perpetuates harmful stereotypes that echo historical anti-Semitic propaganda like "The Protocols of the Elders of Zion." Instead of fostering division, we should advocate for a more nuanced understanding of the complex dynamics in the film industry and promote equal rights and representation for all communities.
\end{mdframed}

\end{document}